%% file: main.tex
\documentclass{article}

\usepackage{longcat_style}
\usepackage[export]{adjustbox}
\usepackage[utf8]{inputenc} %
\usepackage[T1]{fontenc}    %
\usepackage{newunicodechar}
\usepackage{hyperref}       %
\usepackage{xcolor}
\usepackage[normalem]{ulem} %
\hypersetup{
    colorlinks=true,      %
    linkcolor=blue,      %
    urlcolor=blue,       %
    citecolor=blue,      %
    linkbordercolor=blue, %
    urlbordercolor=blue,
    citebordercolor=blue,
    pdfborderstyle={/S/U/W 1}, %
}

\usepackage{float}
\usepackage{url}            %
\usepackage{booktabs}       %
\usepackage{amsfonts}       %
\usepackage{nicefrac}       %
\usepackage{microtype}      %
\usepackage{lipsum}		%
\usepackage{graphicx}
\usepackage{natbib}
\usepackage{doi}
\usepackage{amsmath}
\usepackage{amssymb} %
\usepackage{xspace}
\usepackage{enumitem}
\usepackage{multirow}
\usepackage{subcaption} 
\usepackage{makecell}
\usepackage{hyperref, cleveref}
\usepackage{pifont}
\usepackage[inkscapelatex=false]{svg}
\usepackage{caption}
\usepackage{wrapfig}
\usepackage{algorithm}
\usepackage{algorithmic}
\usepackage{threeparttable}
\usepackage{overpic}

\setlist[itemize]{leftmargin=*}
\setlist[enumerate]{leftmargin=*}
\setlist[description]{leftmargin=*}

\newcommand{\longcat}{LongCat-Video\xspace}

\definecolor{midnightgreen}{rgb}{0.0, 0.29, 0.33}

\title{\textnormal{\longcat Technical Report}}

\author{
    Meituan LongCat Team
}

\renewcommand{\headeright}{\raisebox{-0.2\height}{\includegraphics[height=1.8em]{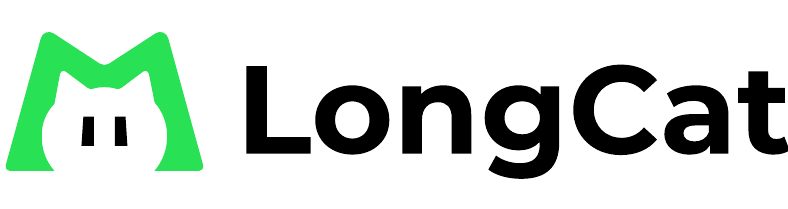}}}
\renewcommand{\shorttitle}{LongCat-Video Technical Report}

\usepackage{amsmath}
\begin{document}
\maketitle

\input{sec/0_abstract}

\clearpage
\tableofcontents
\clearpage

\input{sec/1_introduction}
\input{sec/2_data}
\input{sec/3_method}

\input{sec/4_training}  
\input{sec/5_evaluation}
\input{sec/6_conclusion}
\input{sec/7_contributors}

\bibliographystyle{unsrtnat}
\bibliography{references}  

\clearpage
\appendix

\input{sec/8_appendix}

\end{document}

%% file: sec/0_abstract.tex
\begin{abstract}

Video generation is a critical pathway toward world models, with efficient long video inference as a key capability. Toward this end, we introduce LongCat-Video, a foundational video generation model with 13.6B parameters, delivering strong performance across multiple video generation tasks. It particularly excels in efficient and high-quality long video generation, representing our first step toward world models. Key features include: \textbf{Unified architecture for multiple tasks}: Built on the Diffusion Transformer (DiT) framework, LongCat-Video supports \textit{Text-to-Video}, \textit{Image-to-Video}, and \textit{Video-Continuation} tasks with a single model; \textbf{Long video generation}: Pretraining on \textit{Video-Continuation} tasks enables LongCat-Video to maintain high quality and temporal coherence in the generation of minutes-long videos; \textbf{Efficient inference}: LongCat-Video generates $720p$, $30fps$ videos within minutes by employing a coarse-to-fine generation strategy along both the temporal and spatial axes. Block Sparse Attention further enhances efficiency, particularly at high resolutions; \textbf{Strong performance with multi-reward RLHF}: Multi-reward RLHF training enables LongCat-Video to achieve performance on par with the latest closed-source and leading open-source models. Code and model weights are publicly available to accelerate progress in the field.

\vspace{0.1cm}


\textbf{GitHub}: \href{https://github.com/meituan-longcat/LongCat-Video}{https://github.com/meituan-longcat/LongCat-Video}

\begin{figure}[htbp]
  \centering
  \includegraphics[width=0.9\textwidth]{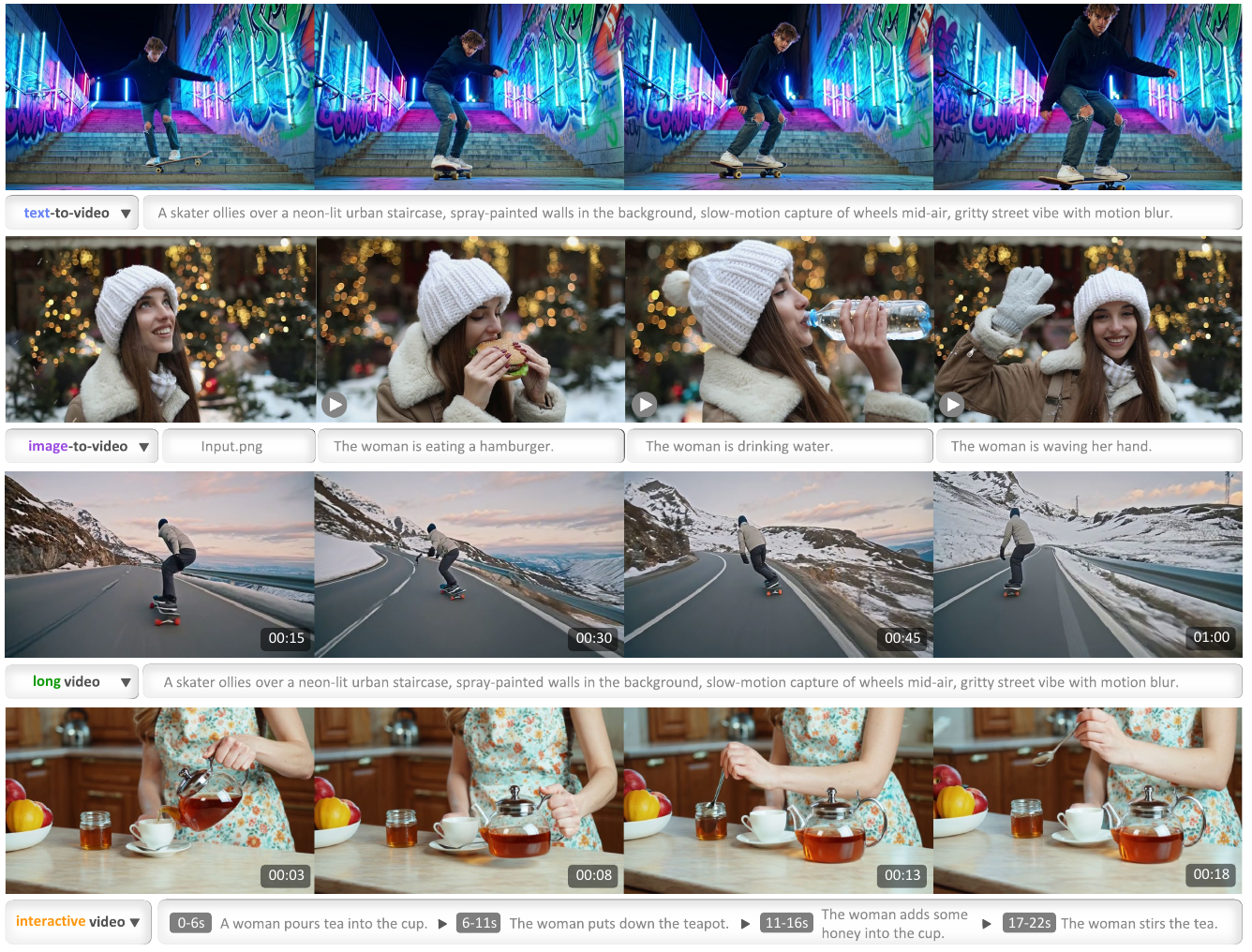}
  \caption{Examples on \textit{Text-to-Video}, \textit{Image-to-Video} and \textit{Video-Continuation} tasks. \textit{Video-Continuation} supports long video generation as well as interactive generation with multiple instructions. We unify these tasks with a single model.}
\end{figure}
    
\end{abstract}

%% file: sec/1_introduction.tex
\section{Introduction}

World models, which aim to understand, simulate, and predict complex real-world environments, constitute an important foundation for applying artificial intelligence in real-world scenarios. Video generation models serve as a critical pathway toward world models by compressing geometric, semantic, physical, and other forms of knowledge through video generation tasks, thereby enabling effective simulation and prediction of the physical world. Notably, efficient long video generation is particularly essential. 

Over the past years, diffusion modeling and video generation have achieved remarkable breakthroughs. The quality of generated videos, instruction-following capabilities, and motion realism have all seen substantial improvements. Commercial products—such as Veo~\citep{veo}, Sora~\citep{sora}, Seedance~\citep{gao2025seedance}, Kling~\citep{Kling}, Hailuo~\citep{Hailuo}, PixVerse~\citep{PixVerse} and others—and open-source solutions—such as Wanx~\citep{wan2025wan}, HunyuanVideo~\citep{kong2024hunyuanvideo}, Step-Video~\citep{ma2025step}, CogVideoX~\citep{yang2024cogvideox} and others—have demonstrated outstanding performance across various dimensions. These works are increasingly being integrated into content production pipelines, with widespread applications ranging from user-generated video content creation to film production, and from entertainment content creation to advertising creativity. Video generation~\citep{nvidia_cosmos} is also establishing a robust foundation for world model applications such as autonomous driving and embodied AI, with the ongoing improvements in physical simulation and long video generation. These developments are further accelerating the deployment and evolution of intelligent systems in complex real-world scenarios.

In this report, we introduce LongCat-Video, a foundational video generation model with 13.6B parameters that delivers strong performance across general video generation tasks, particularly excelling in efficient, high-quality long video generation. LongCat-Video serves as a robust general-purpose model and marks our first step toward world models. Key features include:

\begin{itemize}

\item \textbf{Unified architecture for multiple tasks} Different use cases demand distinct video generation functionalities. For example, \textit{Text-to-Video} is widely adopted for creative content production, while \textit{Image-to-Video} is preferred when precise content control is required. LongCat-Video unifies \textit{Text-to-Video}, \textit{Image-to-Video}, and \textit{Video-Continuation} tasks within a single video generation framework, distinguishing them by the number of conditioning frames—zero for \textit{Text-to-Video}, one for \textit{Image-to-Video}, and multiple for \textit{Video-Continuation} generation. Through a multi-task training strategy, LongCat-Video natively supports all these tasks and delivers strong performance across them.

\item \textbf{Long video generation} Long-video generation is critical for applications such as digital humans, embodied AI, and other complex tasks that require extended temporal coherence, which is also a key capability for world model applications. However, this remains a challenging problem due to generation error accumulated over time. While various methods~\citep{chen2025skyreelsv2infinitelengthfilmgenerative} exist to finetune existing video foundation models for improved long-video generation, LongCat-Video is natively pretrained on \textit{Video-Continuation} tasks, enabling it to produce minutes-long videos without color drifting or quality degradation.

\item \textbf{Efficient inference} The computational cost of video generation increases substantially with higher video resolutions and frame rates, as attention complexity grows quadratically with the number of tokens. Inspired by Seedance~\citep{gao2025seedance}, Hailuo~\citep{Hailuo} and related works, LongCat-Video adopts a coarse-to-fine strategy: videos are first generated at $480p, 15 fps$, and subsequently refined to $720p, 30 fps$. For high-resolution generation, we train an expert LoRA module to effectively leverage the base model’s knowledge. Furthermore, we implement a block sparse attention mechanism, reducing attention computations to less than $10\%$ of those required by standard dense attention. This design significantly enhances efficiency in the high-resolution refinement stage.
  
\item \textbf{Strong performance with multi-reward RLHF} In post-training, we employ Group Relative Policy Optimization (GRPO)~\citep{guo2025deepseek_r1} method to further enhance model performance using multiple rewards. Comprehensive evaluations on both internal and public benchmarks, using human and model-based annotations, demonstrate that LongCat-Video achieves performance comparable to leading open-source video generation models as well as the latest commercial solutions. We are releasing the code, model weights, and key modules, including block sparse attention, to the community. We believe this work will help advance the development of video generation technology in both academic and industrial domains.
  
\end{itemize}

%% file: sec/2_data.tex
\section{Data}

\begin{figure}[htbp]
    \centering
    \includegraphics[width=1\linewidth]{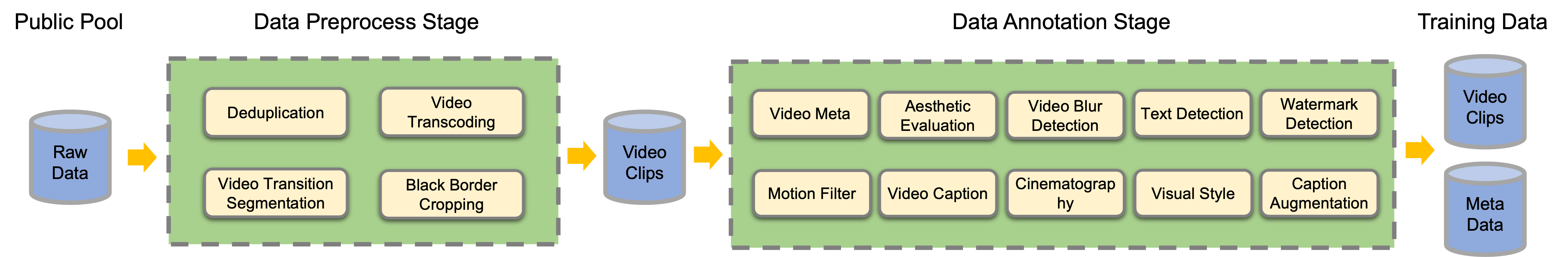}
    \caption{Overview of data curation pipeline. The data preprocessing stage extracts well-segmented video clips from raw source videos in the data pool. In the data annotation stage, each video clip is annotated with a variety of attributes, forming a comprehensive metadata database. This metadata database enables the convenient and flexible assembly of training datasets to support various training stages and objectives.}
    \label{fig:DataCurationPipeline}
\end{figure}

Training a high-quality video generation model requires a large-scale, diverse, and high-quality dataset. To meet these requirements, we have developed a comprehensive data curation pipeline, as illustrated in Figure~\ref{fig:DataCurationPipeline}, which consists of two main stages: 1) \textbf{Data Preprocessing Stage}: This stage includes the acquisition of various data sources, deduplication, video transition segmentation, and black border cropping, ensuring the diversity and integrity of the collected videos; 2) \textbf{Data Annotation Stage}: In this stage, video clips are annotated with multiple metrics and attributes to enrich the dataset and facilitate downstream tasks. We introduce the data curation pipeline in Section~\ref{sec:datacurationpipeline} and present the distribution of the curated training data in Section~\ref{sec:datadistribution}.
 
\subsection{Data Curation Pipeline}
\label{sec:datacurationpipeline}

\subsubsection{Data Preprocessing Stage}

We collect raw video data from a variety of sources. To eliminate redundant content, we perform deduplication using source video IDs and MD5 hashes. PySceneDetect~\citep{Castellano_PySceneDetect} and an in-house trained TransNetV2~\citep{souček2020transnetv2effectivedeep} are employed to segment source videos into training-friendly clips while maintaining content consistency within each fragment—an essential factor for effective video generation model training. Additionally, black border cropping is applied using FFMPEG~\citep{ffmpeg} during the video transition segmentation process to further improve data quality. Finally, all processed video clips are compressed and packaged, facilitating subsequent data cleaning and efficient data loading during training.

\subsubsection{Data Annotation Stage}

To meet the video filtering requirements at different training stages, we annotate video clips with a range of metrics and store them as a comprehensive metadata library. These metrics include basic video metadata (such as duration, resolution, frame rate, and bitrate), aesthetic score, blur score, text coverage, watermark detection, etc. Additionally, motion information is evaluated using extracted video optical flow to assess video dynamics, enabling us to filter out clips with minimal motion features. This metadata library facilitates flexible and targeted dataset construction for various training objectives.

\begin{figure}[htbp]
    \centering
    \includegraphics[width=1.00\textwidth]{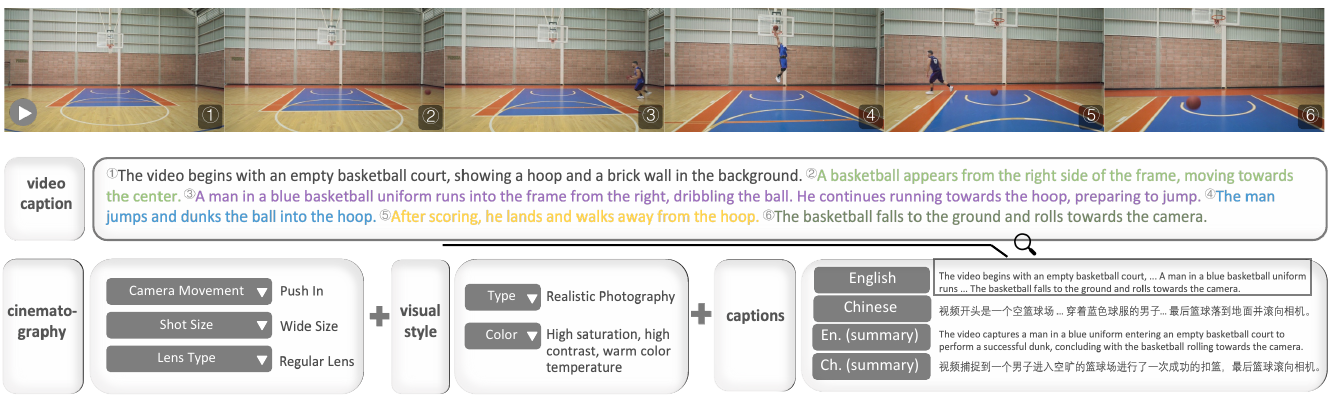}
    \caption{Overview of the video captioning workflow. The main content of each video is captured by a basic captioning model, and complemented by additional models that extract attributes such as cinematography and visual style. These elements are integrated to produce varied and informative captions, enhancing the quality and diversity of training data.}
    \label{fig:caption_example}
\end{figure}

The consistency between captions and video content is crucial for ensuring that the video generation model can accurately follow instructions. As illustrated in Figure~\ref{fig:caption_example}, we decompose the video information and utilize multiple models to annotate various aspects of the video content.

\paragraph{Basic video caption} Videos contain complex information, including both appearance features and the temporal dynamics of actions and events. Many multimodal models are good at describing static images, but struggle to accurately capture actions and understand temporal relationships. We fine-tune the LLaVA-Video model~\citep{zhang2024video} using in-house constructed synthetic video-text pairs, improving its ability to describe both visual and temporal aspects. We also found that the amount and quality of temporal action annotations in the dataset are key to enhancing temporal understanding. To further improve this, we collected more videos with rich temporal events and used annotated data from Tarsier2~\citep{yuan2025tarsier2} for fine-tuning. This significantly boosts the model’s ability to describe and understand temporal dynamics in videos.

\paragraph{Cinematography and visual style} Cinematography in video includes elements such as camera movements, shot sizes, and lens types. To enable automatic recognition of camera movements, we annotated a dataset with categories including pan, tilt, zoom, and shark, and trained a dedicated classifier. The annotation of shot sizes and lens types requires image-level semantic understanding; for this purpose, we employ the Qwen2.5VL model~\citep{bai2025qwen2}, which excels at image analysis and accurately identifies these attributes. Visual style covers a broad range of characteristics, including general visual types such as realism, 2D anime, and 3D cartoon, as well as finer-grained attributes like color tones. For visual style annotation, we likewise utilize Qwen2.5VL, leveraging its strong image understanding capabilities to capture and interpret these diverse visual features.

\paragraph{Caption augmentation} To improve the model's robustness in handling diverse textual inputs, we enrich video captions through a variety of augmentation techniques. These include translating captions between Chinese and English to support both languages, as well as generating concise summaries to diversify caption styles. As illustrated in Figure~\ref{fig:caption_example}, we further enhance caption diversity by randomly selecting elements from cinematography and visual style categories and integrating them with the augmented captions. This strategy ensures that each video clip is paired with multiple styles of textual descriptions, significantly increasing dataset diversity and enhancing the adaptability of the video generation model.

\subsection{Data Distribution}
\label{sec:datadistribution}

\begin{figure}[htbp]
    \centering
    \includegraphics[width=1.0\linewidth]{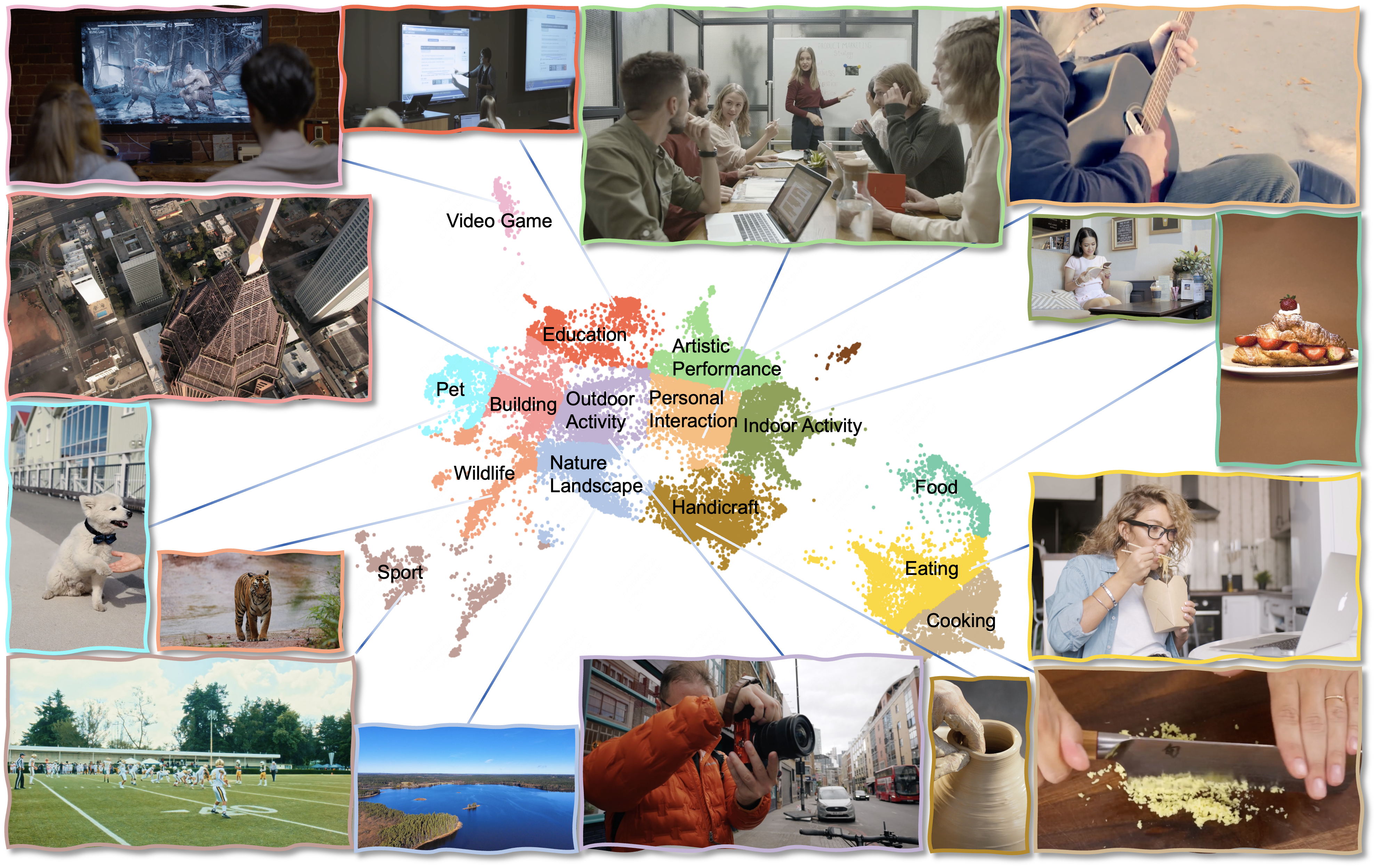}
    \caption{We apply text embedding  to video captions and perform clustering analysis. An LLM summarizes each cluster and assigns tags, enabling unsupervised categorization of the dataset.}
    \label{fig:distribution}
\end{figure}

As shown in Figure~\ref{fig:distribution}, we categorize video clips into several content types by performing cluster analysis on text embedding vectors derived from their captions. (e.g., personal interactions, artistic performances, natural landscapes, etc.). We then assess the data volume and distribution density for each category to evaluate the overall uniformity of the dataset. Based on this analysis, we implement targeted data supplementation or rebalancing strategies as needed. This systematic approach allows for dynamic and precise allocation of data subsets tailored to the specific requirements and objectives of different training phases, thereby optimizing the model training workflow.



%% file: sec/3_method.tex
\section{Method}

\subsection{Model Architecture}

\paragraph{Network Architecture} We employ a standard DiT~\citep{peebles2023scalable} architecture with single-stream transformer blocks. Each block consists of a 3D self-attention layer, a cross-attention layer for text conditioning, and a Feed-Forward Network (FFN) with SwiGLU~\citep{shazeer2020glu}. For modulation, we utilize AdaLN-Zero~\citep{peebles2023scalable}, where each block incorporating a dedicated modulation MLP. To enhance training stability, RMSNorm~\citep{zhang2019root} is applied as QKNorm~\citep{henry2020query} within both the self-attention and cross-attention modules. Additionally, 3D RoPE~\citep{su2024roformer} is adopted for positional encoding of visual tokens. Detailed model specifications are summarized in Table~\ref{tab:networkarchitecture}.

\begin{table}[htbp]
\centering
\caption{Model specifications of LongCat-Video.}
\label{tab:networkarchitecture}
\begin{tabular}{c|c|c|c|c}
\Xhline{1.5pt}
Num. of Layers & Model Hidden Size & FFN Hidden Size & Num. of Attn. Heads & AdaLN Embedding Size \\
\hline
48 & 4096 & 16384 & 32 & 512 \\
\Xhline{1.5pt}
\end{tabular}
\end{table}


\paragraph{VAE and Text embedder} For latent compression, we employ WAN2.1 VAE~\citep{wan2025wan} to convert video pixels into latent tokens, achieving a compression ratio of $4\times8\times8$ along the temporal, height, and width dimensions. In addition, a patchify operation within the DiT model further compresses the latents with an additional $1\times2\times2$ ratio. As a result, the overall compression ratio from pixels to latents reaches $4\times16\times16$. For text encoding, we utilize umT5~\citep{chung2023unimax}, a multilingual text encoder that supports both English and Chinese captions.

\subsection{Unified Model for Multiple Tasks}

\begin{figure}[htbp]
    \centering
    \includegraphics[width=1.0\textwidth]{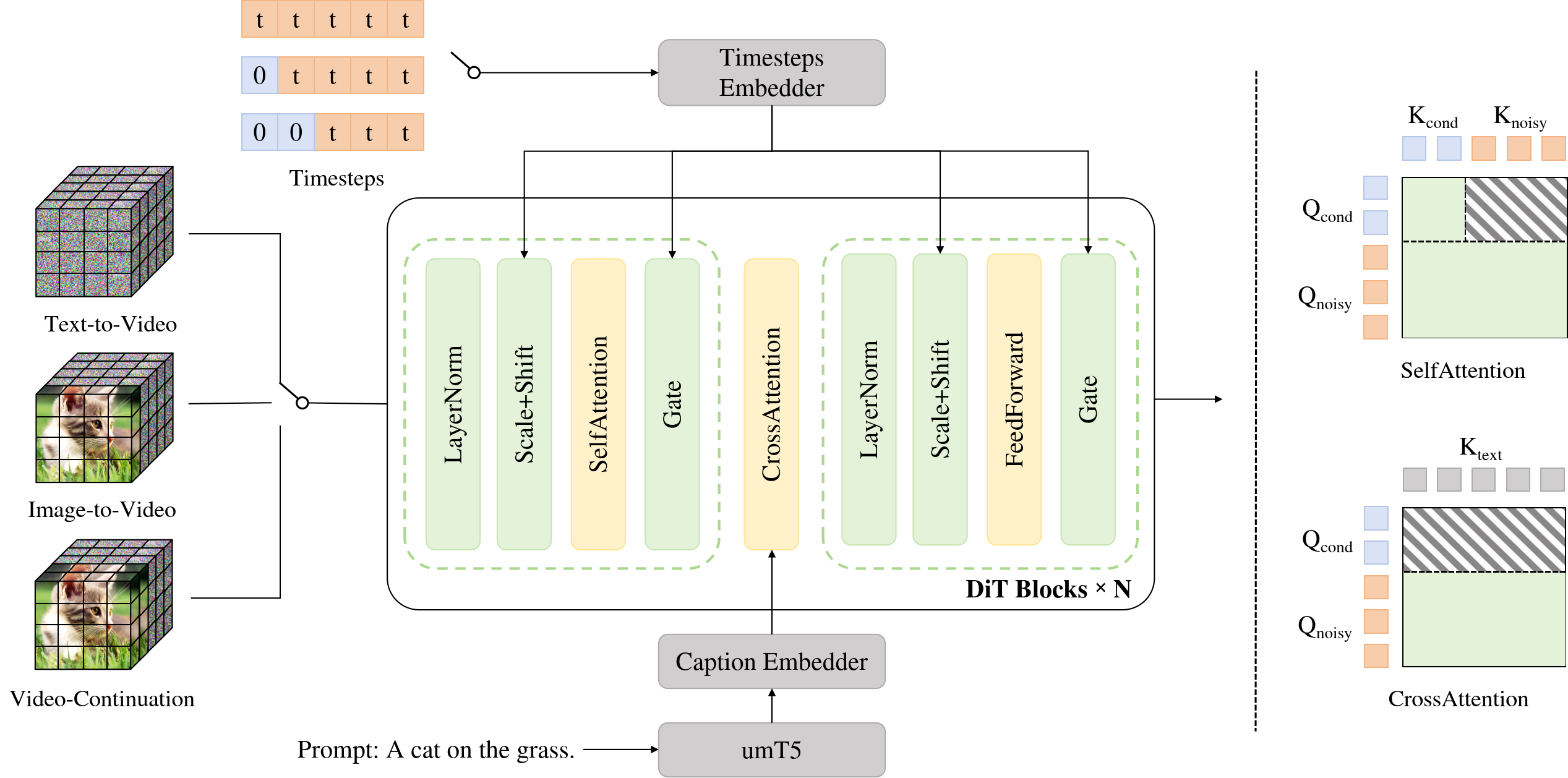}
    \caption{Left: Unified transformer for multiple generation tasks. Our model simultaneously supports \textit{Text-to-Video}, \textit{Image-to-Video} (with a single conditioning frame), and \textit{Video-Continuation} (with multiple conditioning frames) tasks. The timestep configuration is consistent with the input, and the condition part are fixed to zero. Right: Block Causal Attention. In self-attention, the updates of the condition tokens are independent of the noisy tokens. In cross-attention, condition tokens do not participate in cross-attention computation.}
    \label{fig:omni_architecture}
\end{figure}

LongCat-Video is a unified video generation framework that supports \textit{Text-to-Video}, \textit{Image-to-Video}, and \textit{Video-Continuation} tasks. We define all these tasks as video continuation, where the model predicts future frames conditioned on a given set of preceding condition frames. The primary difference between all these tasks is the number of condition frames provided, resulting in a hybrid input format for our network.

\paragraph{Unified Input Representation} 

As illustrated in Figure~\ref{fig:omni_architecture}, the network input consists of two sequences: the condition sequence $X_\text{cond} \in \mathbb{R}^{B \times N_\text{cond} \times H \times W \times C}$, which is the noise-free condition frames, and the noisy sequence $X_\text{noisy} \in \mathbb{R}^{B \times N_\text{noisy} \times H \times W \times C}$, which is the noisy frames to be denoised. Here, $N_\text{cond}$ and $N_\text{noisy}$ denote the lengths of the condition and noisy frames. $B$ is the batch size, $H$ and $W$ are the spatial dimensions, and $C$ is the number of channels. These two sequences are concatenated along the temporal axis to form the overall model input $X \in \mathbb{R}^{B \times (N_\text{cond} + N_\text{noisy}) \times H \times W \times C}$, expressed as $X = [X_\text{cond}, X_\text{noisy}]$ where $[\cdot]$ denotes the concatenation operation. 

Similarly, the timesteps $t$ are partitioned as $t = [t_\text{cond}, t_\text{noisy}]$, where $t_\text{cond}$ corresponds to the timesteps of the condition frames and $t_\text{noisy}$ to those of the noisy frames. This configuration of input sequences and timesteps enables the model to identify different task types based on input patterns. By explicitly structuring both the data and the associated timesteps, the model can effectively distinguish between various generation modes, thereby enhancing its flexibility and performance across a range of generative tasks. For the condition frames, we set $t_\text{cond}$ to $0$ to inject clear, lossless information, while $t_\text{noisy}$ is sampled within the range $[0, 1]$. During loss computation, the contribution from the condition frames is omitted. The condition sequence remains fixed throughout both training and inference.

\paragraph{Block Attention with KVCache} To accommodate the previously described input representation, we have designed a specialized attention mechanism within the unified model architecture, formulated as follows:
\begin{align}
    X_\text{cond} &= \mathrm{Attention}(Q_\text{cond}, K_\text{cond}, V_\text{cond}), \\
    X_\text{noisy} &= \mathrm{Attention}(Q_\text{noisy}, [K_\text{cond}, K_\text{noisy}], [V_\text{cond}, V_\text{noisy}]),
\end{align}
where $Q_\text{cond}$, $K_\text{cond}$, and $V_\text{cond}$ denote the query, key, and value of the condition tokens, and $Q_\text{noisy}$, $K_\text{noisy}$, and $V_\text{noisy}$ correspond to those of the noisy tokens. This design ensures that the condition tokens are not influenced by the noisy tokens. Additionally, $X_\text{cond}$ does not participate in the cross-attention computation. The computation related to condition tokens depends solely on the input video condition frames, allowing us to cache the KV features of the condition tokens and reuse them across all sampling steps, while ensuring consistency between training and inference. This strategy further enhances the efficiency of long video generation. 

\subsection{Multi-Reward GRPO Training}
\label{sec:method_grpo}

\subsubsection{GRPO for Flow Matching Modeling}


Although GRPO has achieved notable success in large language models~\citep{guo2025deepseek_r1} and image generation~\citep{liu2025flow,xue2025dancegrpo,li2025mixgrpo,he2025tempflowgrpotimingmattersgrpo}, its application to video generation is particularly challenging due to slow convergence and complex reward optimization. To overcome these issues, we introduce a series of techniques that significantly enhance both convergence speed and generation quality (Fig. \ref{fig:grpo_cp}) of GRPO for video generation tasks. The theoretical framework is outlined in Appendix~\ref{Apdx:grpo-prelim}, and the complete GRPO training procedure is summarized in Algorithm~\ref{alg:grpo}.

\begin{figure}[thbp]
  \centering
  \includegraphics[width=1\textwidth]{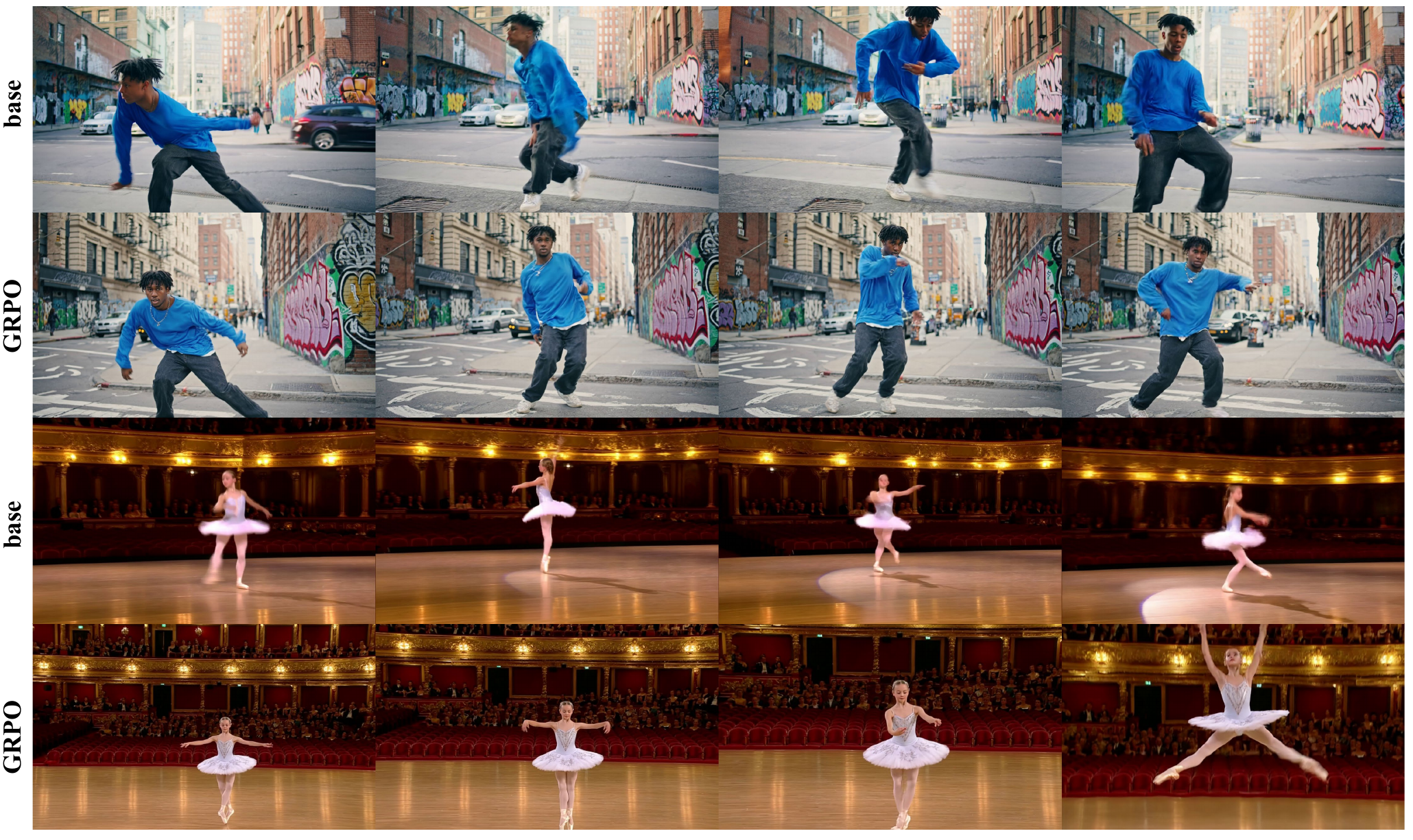}
  \caption{Our GRPO method significantly improves the video generation quality.}
  \label{fig:grpo_cp}
\end{figure}

\begin{algorithm}
\caption{LongCat-Video's GRPO Training for Flow Matching Models}
\label{alg:grpo}
\begin{algorithmic}[1]
\REQUIRE Prompt distribution $\mathcal{C}$, group size $G$, total timesteps $T$, reward models $\{R_k\}_{k=1}^n$, weights $\{w_k\}_{k=1}^n$
\ENSURE Optimized policy parameters $\theta$
\STATE Initialize policy parameters $\theta$, reference policy $\pi_{\text{ref}}$
\REPEAT
\STATE Sample batch of prompts $\{c_j\}_{j=1}^B \sim \mathcal{C}$
\FOR{each prompt $c_j$ \textbf{in parallel}}
    \STATE \textcolor{blue}{// Fix the initial noise and SDE timestep (Sec. \ref{sec:grpo-fixsde})}
    \STATE Sample initial noise $\boldsymbol{x}_T \sim \mathcal{N}(0, I)$
    \STATE Sample critical timestep $t' \sim \mathcal{U}(0, T'-1)$
    \FOR{$i = 1$ \TO $G$}
        \STATE Generate trajectory $\{\boldsymbol{x}_t^i\}_{t=0}^T$:
        \FOR{$t = T$ \TO $0$}
            \IF{$t = t'$}
                \STATE $\boldsymbol{x}_{t-1}^i \leftarrow \boldsymbol{x}_t^i + \text{drift}_{\theta}(\boldsymbol{x}_t^i, t, c_j)\Delta t + \sigma_t\sqrt{\Delta t}\epsilon$ \textcolor{blue}{// SDE step with truncated noise schedule (Sec. \ref{sec:grpo-trunc})}
            \ELSE
                \STATE $\boldsymbol{x}_{t-1}^i \leftarrow \boldsymbol{x}_t^i + \text{drift}_{\theta}(\boldsymbol{x}_t^i, t, c_j)\Delta t$ \textcolor{blue}{// ODE step}
            \ENDIF
        \ENDFOR
        \STATE Compute rewards $\{R_k(\boldsymbol{x}_0^i, c_j)\}_{k=1}^n$
    \ENDFOR
    \FOR{$k = 1$ \TO $n$}
        \STATE Compute $\mu_k \leftarrow \text{mean}(\{R_k(\boldsymbol{x}_0^i, c_j)\}_{i=1}^G)$
        \STATE Compute $\sigma_k^j \leftarrow \text{std}(\{R_k(\boldsymbol{x}_0^i, c_j)\}_{i=1}^G)$
        \STATE Collect $\{\sigma_k^j\}_{j=1}^B$ from all processes
        \STATE Compute $\sigma_{max, k} \leftarrow \text{max}(\{\sigma_k^j\}_{j=1}^B)$ \textcolor{blue}{// max group std (Sec. \ref{sec:grpo-maxstd})}
        \FOR{$i = 1$ \TO $G$}
            \STATE $\hat{A}_{k,t'}^i \leftarrow \frac{R_k(\boldsymbol{x}_0^i, c_j) - \mu_k}{\sigma_{max, k}}$ 
        \ENDFOR
    \ENDFOR
    \FOR{$i = 1$ \TO $G$}
        \STATE \textcolor{blue}{// Weighted relative advantage for multi-reward (Sec. \ref{sec:grpo-multireward})}
        \STATE $\hat{A}_{\text{total}}^i \leftarrow \sum_{k=1}^n w_k \hat{A}_{k,t'}^i$ \\
        \STATE \textcolor{blue}{// Reweighting of the Policy and KL Loss (Sec. \ref{sec:grpo-rew})}
        \STATE $\lambda_{\text{policy}} \leftarrow \sqrt{\frac{\frac{t'}{T}}{\Delta \frac{t'}{T} (1-\frac{t'}{T})}}$
        \STATE $\lambda_{\text{KL}} \leftarrow \frac{t'}{\Delta \frac{t'}{T} (1-\frac{t'}{T})}$
        \STATE $\mathcal{L}_{\text{policy}}^i \leftarrow \lambda_{\text{policy}} \cdot r_{t'}^i(\theta) \cdot \hat{A}_{\text{total}}^i$
        \STATE $\mathcal{L}_{\text{KL}}^i \leftarrow \beta \lambda_{\text{KL}} \cdot D_{\mathrm{KL}}(\pi_\theta \| \pi_{\text{ref}})$
        \STATE $\mathcal{L}^i \leftarrow \mathcal{L}_{\text{policy}}^i - \mathcal{L}_{\text{KL}}^i$
    \ENDFOR
\ENDFOR
\STATE $\mathcal{L}_{\text{total}} \leftarrow \frac{1}{B \cdot G} \sum_{j=1}^B \sum_{i=1}^G \mathcal{L}^i$
\STATE $\theta \leftarrow \theta - \eta \nabla_\theta \mathcal{L}_{\text{total}}$
\UNTIL{convergence}
\end{algorithmic}
\end{algorithm}

\paragraph{GRPO as stochastic noise search} We observe that GRPO for Flow Matching~\citep{lipman2022flow} effectively simulates the gradients $\frac{d R}{d v_\theta}$ using stochastic noise search. In our reweighted version of the policy loss (See Appendix \ref{Apdx:grpo-rew} for details.), the gradient of the policy loss with respect to the model parameter $\theta$ is as follows:

\begin{equation}
\label{eq:policy_reweight_zhengwen}
\nabla_\theta \mathcal{L}_{\text{policy, reweighted}}(\theta) = -\frac{3}{2} \hat{A}_t^i  \cdot \epsilon \cdot \nabla_\theta v_\theta
\end{equation}

It is worth noting that Eq.(\ref{eq:policy_reweight_zhengwen}) reveals that in flow matching models, GRPO fundamentally uses the relative advantage $\hat{A}_t^i$ and the noise term $\epsilon$ in the stochastic differential equation (SDE) sampling~\citep{song2020score} to approximate $\frac{d R}{d v_\theta}$, the gradient of the reward with respect to the velocity field, following the chain rule decomposition:

\begin{equation}
\frac{d R}{d \theta} = \frac{d R}{d v_\theta} \cdot \frac{d v_\theta}{d \theta}
\end{equation}

where the GRPO framework provides the specific form:

\begin{equation}
\frac{d R}{d v_\theta} \approx -\frac{3}{2} \hat{A}_t^i \cdot \epsilon
\end{equation}

Based on this finding, we design the following strategies.

\paragraph{Fix the stochastic timestep in SDE sampling}
\label{sec:grpo-fixsde}

Previous GRPO methods for Flow Matching sample trajectories using SDE sampling at all timesteps. This approach introduces temporal credit assignment ambiguity, as the reward is not accurately attributed to the specific timesteps that contributed to the final outcome. Instead, the reward is uniformly distributed across all timesteps, including those that may not have made a positive contribution. To address this ambiguity, we introduce a modified sampling scheme that isolates reward variation. Similar to concurrent works~\citep{he2025tempflowgrpotimingmattersgrpo, zhou2025g2rpogranulargrpoprecise}, for each prompt $c$, samples share the same initial noise latent, and a single critical timestep $t$ is randomly selected from the first $T^{\prime}$ timesteps ($T^{\prime} < T$). SDE sampling with noise injection is applied only at $t$, while all other timesteps use deterministic ordinary differential equation (ODE) sampling. This approach enables precise credit assignment and leads to more stable, interpretable policy optimization. See Appendix~\ref{Apdx:grpo-fix-sde} for details.

\paragraph{Truncated noise schedule}
\label{sec:grpo-trunc}

To enhance the diversity of SDE sampling, we adopt an amplified noise schedule with coefficient $a = 1$. However, this aggressive schedule can cause instability at high noise levels, as the diffusion coefficient $\sigma_t \sqrt{\Delta t}$ becomes excessively large when $t$ approaches $1$. We introduce a threshold-based clipping mechanism for the diffusion term. Specifically, the diffusion coefficient is clipped when it exceeds a predefined threshold $\tau$:
\[
\sigma_t \sqrt{\Delta t} \rightarrow \min\left(\sigma_t \sqrt{\Delta t}, \tau\right).
\]


When clipping occurs, we set $\sigma_t$ in the drift term to $\tau / \sqrt{\Delta t}$ for consistency. In our experiments, $\tau$ is set to $0.45$.



\paragraph{Policy and KL Loss reweighting}
\label{sec:grpo-rew}

The gradient of the policy loss with respect to $\theta$ is as follows (See Appendix \ref{Apdx:grpo-rew} for details):

\begin{equation}
\nabla_\theta \mathcal{L}_{\text{policy}}(\theta) = -\frac{3}{2} \hat{A}_t^i  \cdot \sqrt{\frac{\Delta t(1-t)}{t}} \cdot \epsilon \cdot \nabla_\theta v_\theta
\end{equation}



We observe that the gradient magnitude is scaled by the factor $\kappa(t, \Delta t) = \sqrt{\frac{\Delta t (1-t)}{t}}$, which introduces two key optimization challenges: 
(1) \textbf{Vanishing gradient:} as $t \rightarrow 1$, $\kappa(t, \Delta t)$ approaches zero, causing the gradient magnitude to vanish in high noise stages; 
(2) \textbf{Small timestep:} video generation models typically use large shifts in timestep scheduling for both training and inference, resulting in small $\Delta t$ values that further suppress the gradient magnitude.

To address these issues, we introduce a reweighting coefficient defined as:

\begin{equation}
\lambda_{\mathrm{policy}}(t, \Delta t) = \kappa(t, \Delta t)^{-1} = \sqrt{\frac{t}{\Delta t(1-t)}}, \quad \mathcal{L}_{\text{policy, reweighted}}(\theta) = \lambda_{\mathrm{policy}}(t, \Delta t) \cdot \mathcal{L}_{\text{policy}}(\theta)
\end{equation}

Similarly, we also introduce a KL reweighting coefficient (See Appendix \ref{Apdx:grpo-rew} for details):

\begin{equation}
\lambda_{\mathrm{KL}}(t, \Delta t) = k_{\mathrm{KL}}(t, \Delta t)^{-1} = \frac{t}{\Delta t(1-t)}, \quad
\mathcal{L}_{\text{KL, reweighted}}(\theta)  = \lambda_{\mathrm{KL}}(t, \Delta t) \cdot D_{\mathrm{KL}}(\theta) 
\end{equation}

\begin{figure}[ht]
\centering
\begin{subfigure}{0.73\textwidth}
  \caption{\hspace*{\fill}} 
  \includegraphics[width=\linewidth]{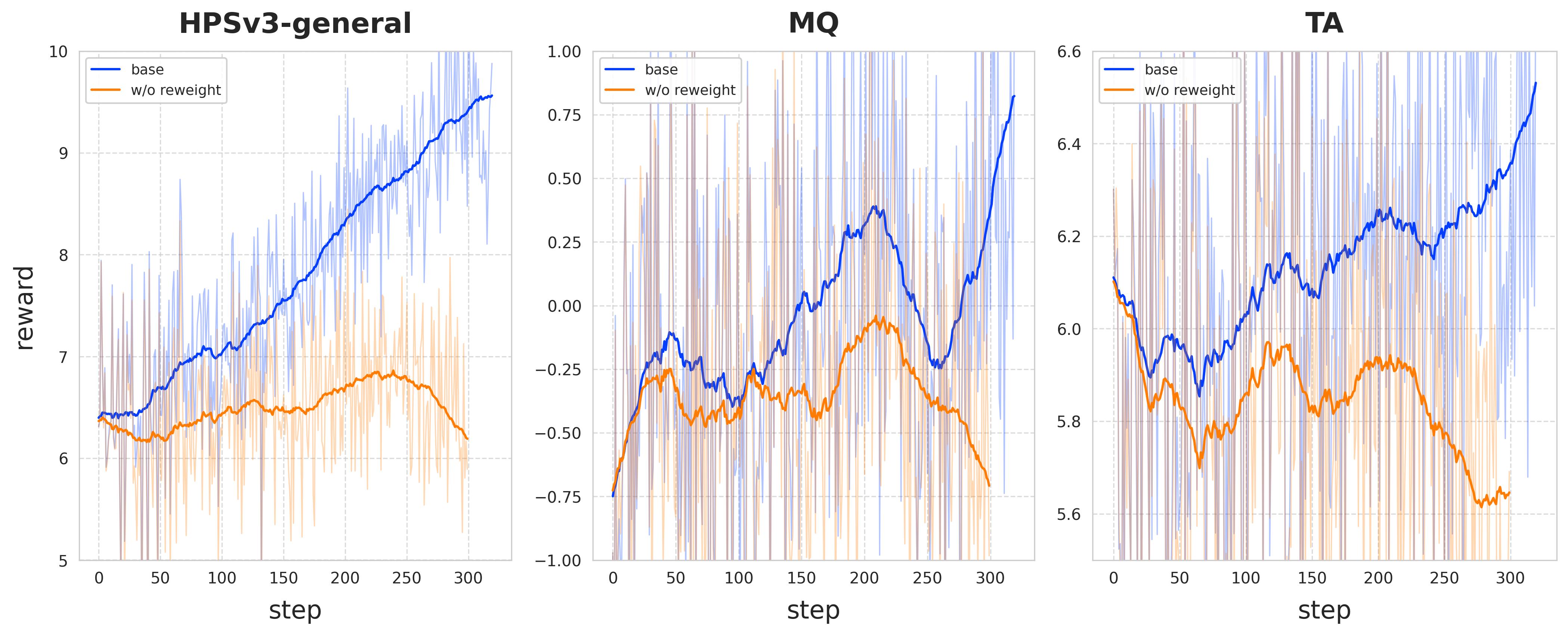}
  \label{fig:ab-reweight}
\end{subfigure}
\hfill
\begin{subfigure}{0.24\textwidth}
  \caption{\hspace*{\fill}} 
  \includegraphics[width=\linewidth]{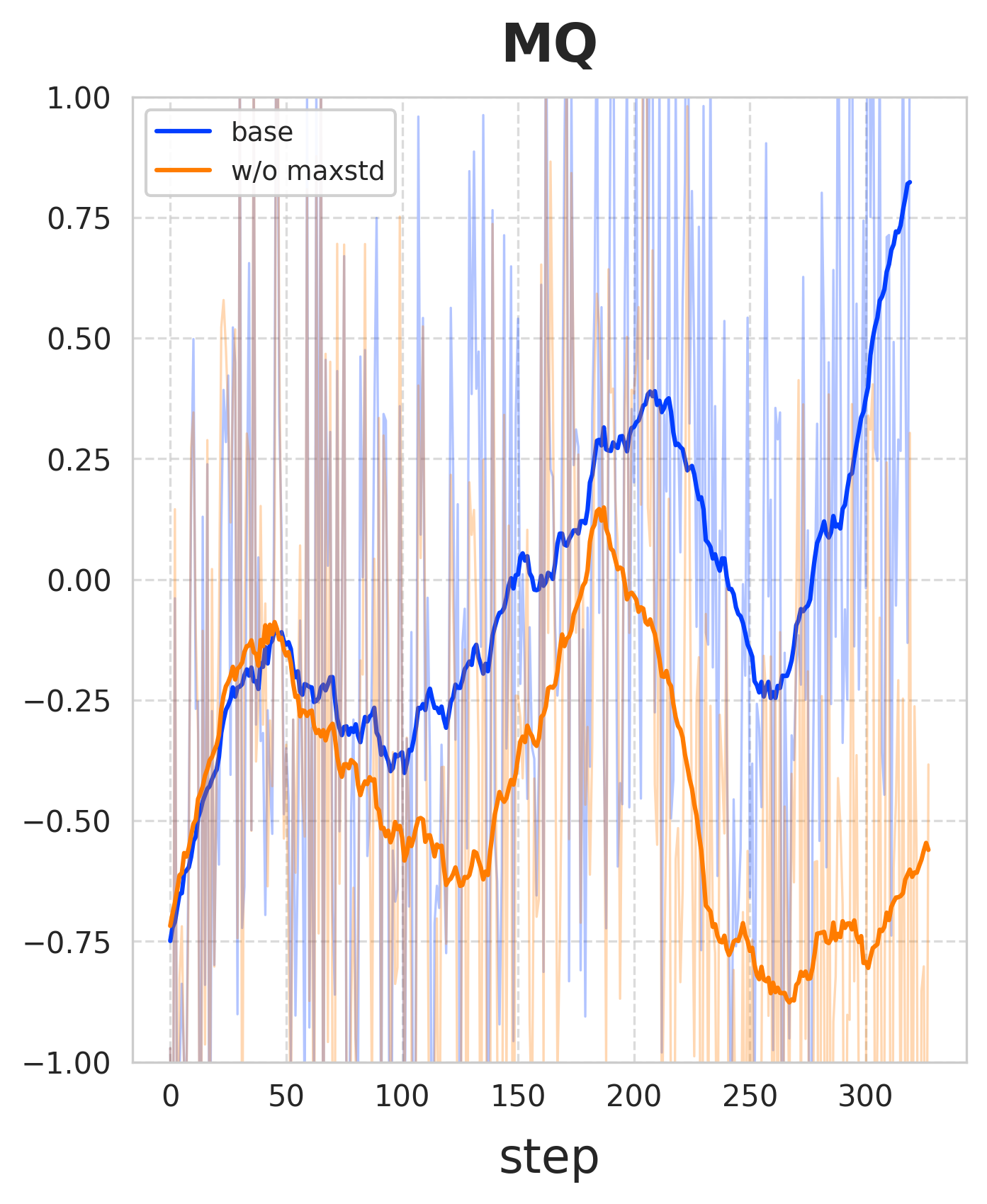}
  \label{fig:ab-maxstd}
\end{subfigure}
\caption{Ablation experiments on: (a) Policy and KL loss reweighting; (b) Max group standard deviation.}
\end{figure}

The reweighting coefficient effectively normalizes the gradient magnitude, eliminating the problematic temporal and step-size dependencies. This ensures stable and efficient optimization throughout the GRPO training (Figure~\ref{fig:ab-reweight}).


\paragraph{Max group standard deviation}
\label{sec:grpo-maxstd}

In the standard GRPO formulation, each prompt corresponds to a group of samples, and the relative advantage is computed using the group-specific standard deviation. However, reward dispersion varies across groups, and those with smaller standard deviations may yield unreliable advantage estimates due to inherent reward model inaccuracies.

To improve training stability, we address this by replacing the group-specific standard deviation with the maximum standard deviation observed across all groups. This adjustment reduces the gradient weight for samples from groups with potentially unreliable advantage estimates, while preserving the signal from groups with more reliable reward distributions. The modified advantage calculation becomes:

\begin{equation}
\hat{A}_{k, t}^i = \frac{R_k\left(\boldsymbol{x}_0^i, c_j\right) - \mu_k}{\sigma_{\max}}
\end{equation}

where $\mu_k$ is the group mean for reward $k$, $\sigma_{\max} = \max_j \sigma_k^j$ is the maximum standard deviation across all groups for reward $k$. This modification ensures that samples from groups with small standard deviations receive appropriately scaled gradient updates and the training process becomes more robust to reward model inaccuracies (Figure~\ref{fig:ab-maxstd}).

\subsubsection{Reward Models and Multi-Reward Training}

\paragraph{Reward Models}

\begin{figure}[thbp]
  \centering
  \includegraphics[width=1\textwidth]{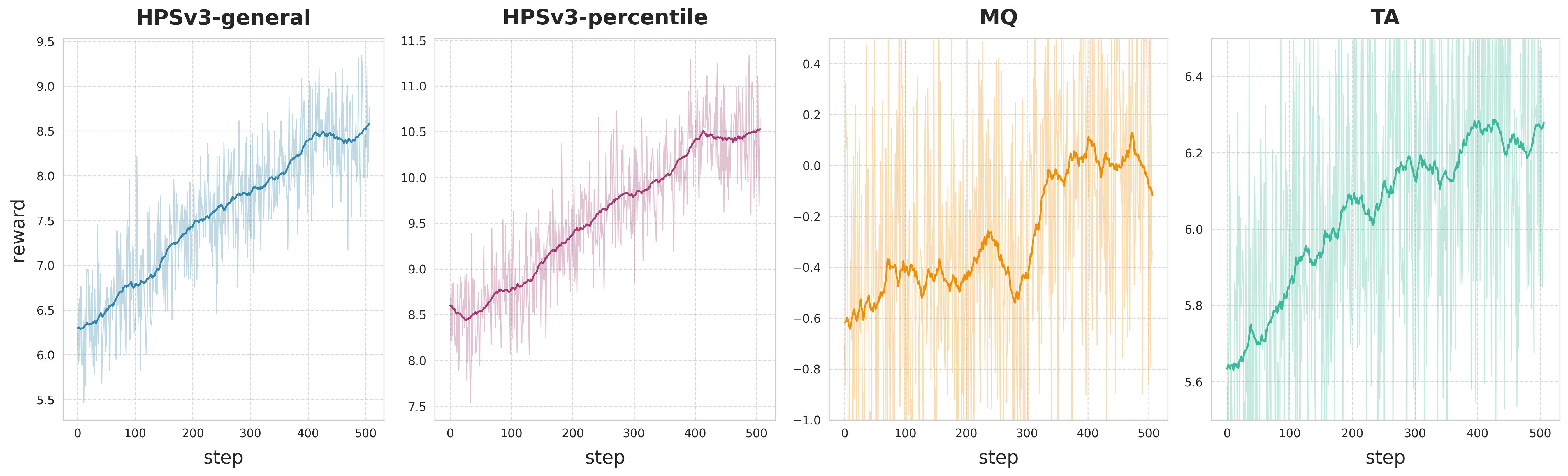}
  \caption{GRPO reward curves from the multi-reward training of LongCat-Video.}
  \label{fig:grpo_rm_cruve}
\end{figure}

\begin{figure}[thbp]
  \centering
  \includegraphics[width=1\textwidth]{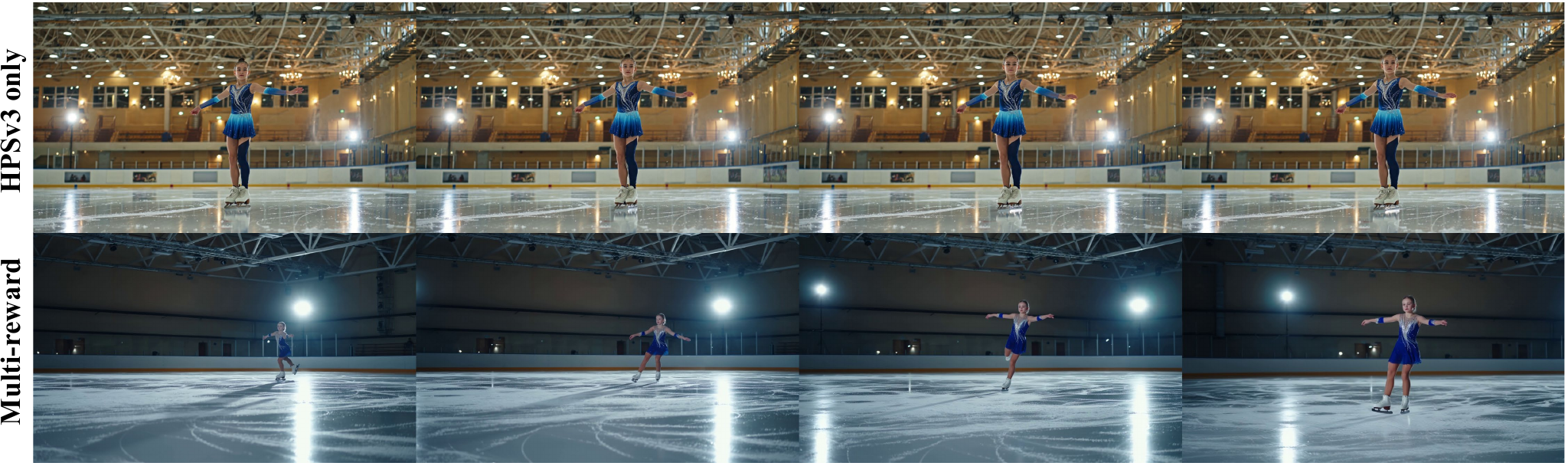}
  \caption{Reward hacking with single reward. Our multi-reward training approach prevents reward hacking for any single reward by establishing a balance among multiple rewards. For instance, the motion reward counteracts the static tendency induced by HPSv3 hacking while still leveraging HPSv3 to enhance visual quality.}
  \label{fig:grpo_hack}
\end{figure}

We utilize three specialized reward models to optimize visual quality (VQ), motion quality (MQ), and text-video alignment (TA) during training.

\begin{itemize}
    \item \textbf{Visual Quality Assessment:} For VQ evaluation, we use HPSv3~\citep{ma2025hpsv3} as our base model, which inherently assesses both visual quality and text-video alignment. We combine two types of HPSv3-based rewards: \textit{\textbf{HPSv3-general}}, which is the mean score of all frames measured with the general prompt \textit{"A high-quality image"} and focuses exclusively on visual quality; and \textit{\textbf{HPSv3-percentile}}, which is measured using the video caption to evaluate text-video alignment and uses the scores of the top 30\% of all frames to mitigate the impact of low rewards resulting from content inconsistency caused by temporal changes.

    \item \textbf{Motion Quality Assessment:} For MQ evaluation, we employ a VideoAlign~\citep{liu2025improving}-based model fine-tuned on internal annotated datasets. To mitigate the model's preference for specific color, we use grayscale videos for both training and inference, which ensures the assessment focuses on motion characteristics rather than color attributes. Additionally, as illustrated in the validation loss curves during training (Figure~\ref{fig:mq_rm_validation_loss}), models trained with grayscale videos show a delayed increase in validation loss compared to those trained with RGB videos, indicating improved generalization and reduced overfitting in MQ reward model training.
    
    \item \textbf{Text-Video Alignment Assessment:} For TA evaluation, we also employ a VideoAlign-based model fine-tuned on internally annotated data. Unlike MQ evaluation, we retain the original color input processing to preserve the model’s ability to assess semantic correspondence between text prompts and video content.

\end{itemize}

\paragraph{Multi-Reward Training}
\label{sec:grpo-multireward}

For multi-reward GRPO training, the effective relative advantage in the policy loss for multi-reward optimization is exactly the weighted sum of the individual relative advantages (Refer to Appendix \ref{Apdx:grpo-multireward} for details). Therefore, the corresponding policy loss becomes:

\begin{equation}
\mathcal{L}_{\text{policy, multi}}(\theta) = r_t^i(\theta) \cdot \left( \sum_{k=1}^n w_k \cdot \hat{A}_{k,t}^i \right)
\end{equation}

where each relative advantage $\hat{A}_{k,t}^i$ is computed independently for reward $R_k$ using group normalization.

In practice, the combination of multiple reward signals provides comprehensive guidance for the policy optimization process, ensuring balanced improvements in all aspects of video generation quality as shown in Figure~\ref{fig:grpo_rm_cruve}. More importantly, the mutual constraints imposed by multiple rewards create a natural regularization effect that prevents over-optimization on any single metric and reduces the likelihood of reward hacking.


\subsection{Efficient Video Generation}
\label{sec:efficient_video_gen}

\begin{figure}[htbp]
    \centering
    \includegraphics[width=1.0\textwidth]{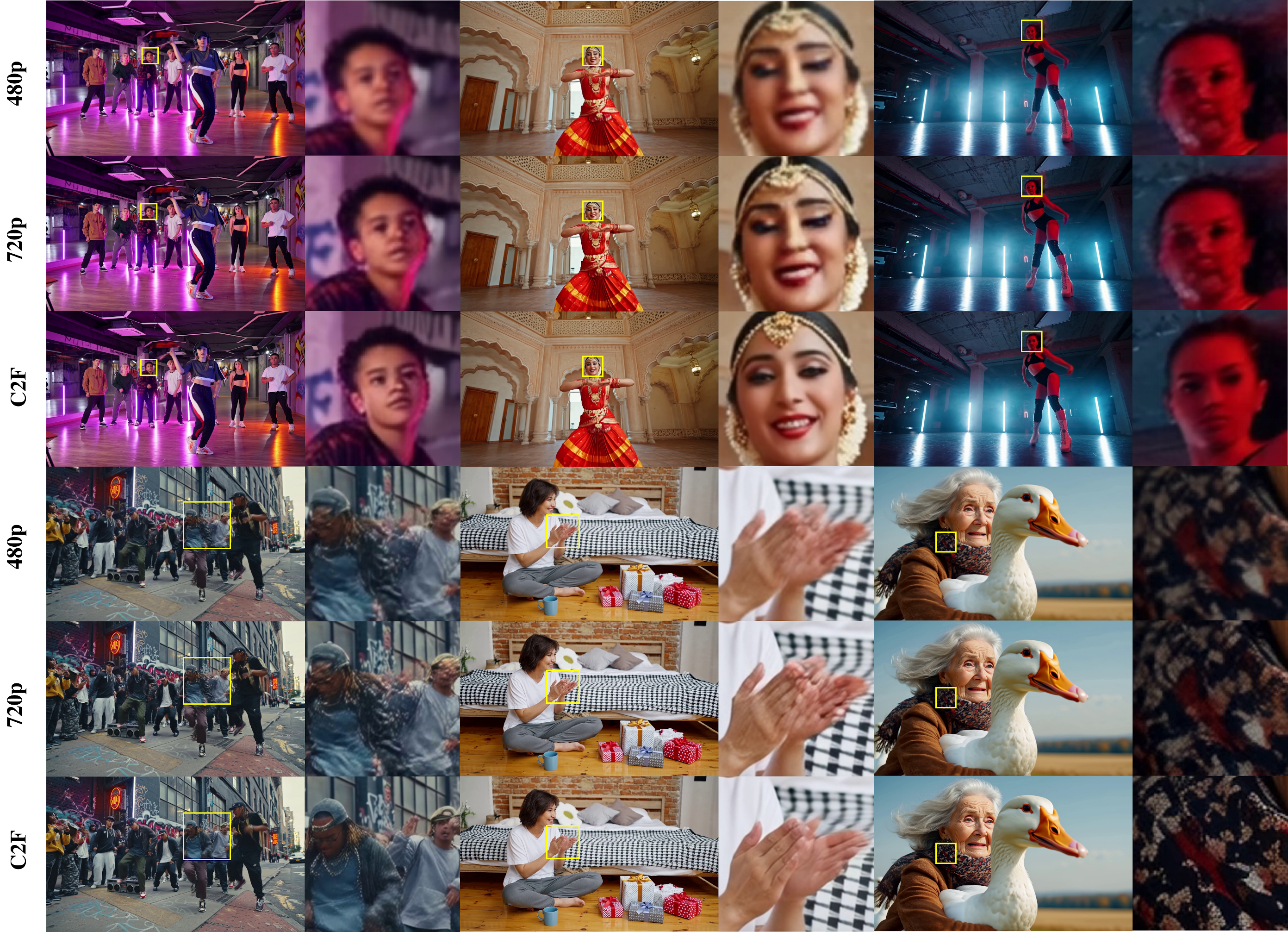}
    \caption{Comparison of native $480p$, native $720p$, and \textit{coarse-to-fine} $720p$ generation. The coarse-to-fine strategy produces texture details and quality that surpass those of the native $720p$ generation and can also correct local distortions.}
    \label{fig:SR}
\end{figure}

Inference efficiency remains a challenge for video generation, particularly for generating high-resolution, high-frame-rate videos. Therefore, we have introduced several optimizations to enhance inference efficiency. We distill the base model to reduce the necessary sampling steps. Additionally, we deploy coarse-to-fine (C2F) generation (Section~\ref{sec:c2f}) and block sparse attention (BSA) (Section~\ref{sec:bsa}) to further reduce the time cost in high-resolution video generation. As shown in Table~\ref{tab:c2fspeedcompare}, combining these strategies increases inference efficiency by more than 10$\times$, allowing $720p, 30 fps$ video generation within minutes. Additionally, we found that the coarse-to-fine generation strategy not only reduces inference cost but also improves generation quality, particularly enhancing visual details, as illustrated in Figure~\ref{fig:SR}.


\begin{table}[htbp]
\centering
\begin{threeparttable}
\caption{Speed comparison under different inference settings.}
\label{tab:c2fspeedcompare}
\begin{tabular}{c|c|c|c|c|c|c}
\Xhline{1.5pt}
Variant & LCM & C2F &BSA & Sampling Steps & Latency & Speedup \\
\hline
$480p \times 93$ frames & \ding{55} & \ding{55} & \ding{55} & 50 & 341.5s & -\\
$480p \times 93$ frames & \ding{51} & \ding{55} & \ding{55} & 16 & 61.3s & -\\
$720p \times 93$ frames & \ding{55} & \ding{55} & \ding{55} & 50 & 1429.5s & 1.0$\times$\\
$720p \times 93$ frames & \ding{51} & \ding{55} & \ding{55} & 16 & 244.6s & 5.8$\times$\\
\hline
$480p \times 93$ frames $\rightarrow$ $720p \times 93$ frames & \ding{51} & \ding{51} & \ding{55} & 16/5 & 135.3s & 10.6$\times$\\
$480p \times 93$ frames $\rightarrow$ $720p \times 189$ frames & \ding{51} & \ding{51} & \ding{55} & 16/5 & 302.9s & 4.7$\times$\\
\hline
$480p \times 93$ frames $\rightarrow$ $720p \times 93$ frames & \ding{51} & \ding{51} & \ding{51} & 16/5 & \textbf{116.5s} & \textbf{12.3$\times$}\\
$480p \times 93$ frames $\rightarrow$ $720p \times 189$ frames & \ding{51} & \ding{51} & \ding{51} & 16/5 & \textbf{142.0s} & \textbf{10.1$\times$}\\
\Xhline{1.5pt}
\end{tabular}
\begin{tablenotes}
  \footnotesize
  \item \hspace{-0.5cm} $\ast$ The tests were conducted on a single H800 GPU with FlashAttention3~\citep{shah2024flashattention3fastaccurateattention}.
\end{tablenotes}
\end{threeparttable}
\end{table}

\subsubsection{Coarse-to-Fine Generation}
\label{sec:c2f}

\begin{figure}[t]
    \centering
    \includegraphics[width=1.0\textwidth]{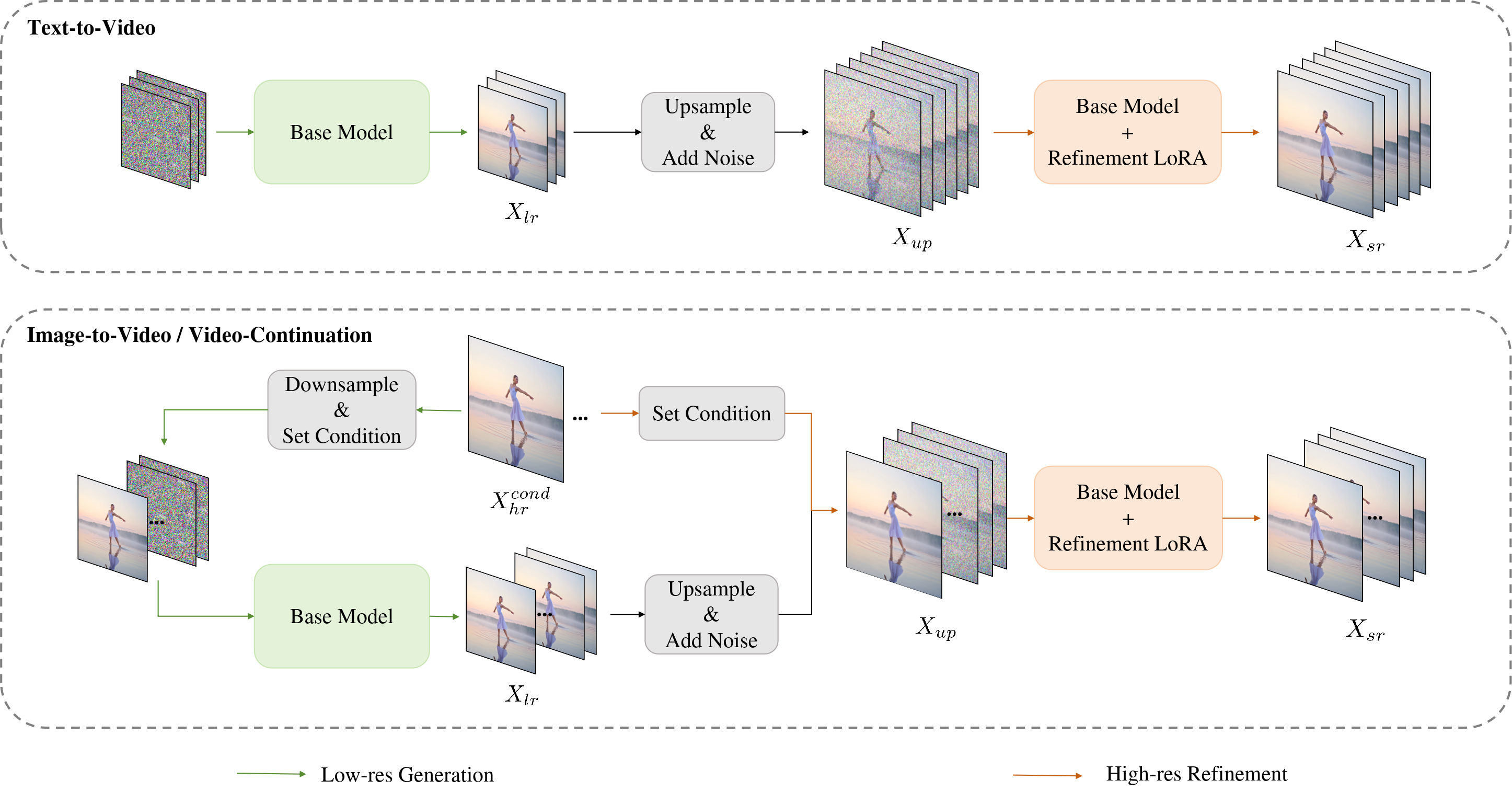}
    \caption{The coarse-to-fine generation processes for \textit{Text-to-Video}, \textit{Image-to-Video}, and \textit{Video-Continuation} tasks. The green arrows indicate the low-resolution generation phase, while the orange arrows represent the refinement phase. Compared to \textit{Text-to-Video}, \textit{Image-to-Video} and \textit{Video-Continuation} include additional configuration for the condition.}
    \label{fig:2stage}
\end{figure}

Training and inference on high-resolution, high-FPS videos incur substantial computational costs due to long token sequences. To address this, we propose a coarse-to-fine generation paradigm (Figure~\ref{fig:2stage}): first, the model generates a $480p, 15 fps$ video; second, this video is upscaled to $720p, 30 fps$ using trilinear interpolation and refined by a refinement expert. This approach greatly improves efficiency and enhances image quality and high-frequency details. The refinement expert is trained with LoRA fine-tuning on the base model. Since the refinement task is similar to the base model's generation task but follows a different denoising path, LoRA enables efficient adaptation while reusing the base model's capabilities. Besides, LoRA fine-tuning is decoupled from other training stages, converges faster, and significantly reduces memory usage.

\paragraph{Refinement using Flow Matching} The training objective of refinement expert is to learn the transformation between the distribution of upsampled $480p, 15 fps$ videos and the distribution of $720p, 30 fps$ videos. We also utilize flow matching to model the mapping between these two distributions. The input to the network for the refinement stage training, denoted as $x_{t'}$, can be represented as follows:

\begin{equation}
x_{t'} = x_0 + (x_{thresh}-x_0)\cdot\frac{t'}{t_{thresh}}, t'\in [0, t_{thresh}],
\end{equation}
\begin{equation}
x_{thresh} = (1-t_{thresh})\cdot x_{up} + t_{thresh}\cdot\epsilon, \epsilon \sim \mathcal{N}(\mathbf{0}, \mathbf{I}),
\end{equation}
\begin{equation}
x_{up} = Encode(Upsample(Decode(x_{lr}))),
\end{equation}

where $x_{lr}$ is the output of the first stage, which is a latent representation of a low-resolution, low-frame-rate video, $x_{up}$ represents the video latent obtained by applying the upsampling operation, denoted as $Upsample$, to $x_{lr}$ in the RGB space, $Encode$ and $Decode$ respectively represent the encoding and decoding processes of the VAE.

To preserve the layout and structural information of low-resolution result, we apply a moderate level of noise, $t_{thresh}$, to $x_{up}$. The result after adding noise is $x_{thresh}$, which serves as the starting point for the refinement stage flow matching path, with the endpoint being $x_0$, the $720p, 30 fps$ video latent. We sample noise intensity $t'$ within the range from 0 to $t_{thresh}$ for training. It should be noted that to ensure the numerical range of the ground truth in the refinement stage aligns with the base model, we need to apply numerical scaling to velocity $x_0-x_{thresh}$. Finally, the ground truth $v_{t'}$ can be expressed as:
\begin{equation}
v_{t'}=\frac{x_0-x_{thresh}}{t_{thresh}}.
\end{equation}
 This design is well-suited to the LoRA training mode, enabling significant reuse of the model's existing knowledge. It is evident that when $t_{thresh}$ is equal to 1, the refinement stage training degenerates into a standard flow matching training process between the standard Gaussian distribution and the high-resolution video distribution. In practice, we set $t_{thresh}$ to 0.5, and the refinement stage requires only 5 sampling steps, significantly improving efficiency. We further combine block sparse attention with the coarse-to-fine generation process, which accelerates sampling even further. Compared to the native generation process of $720p, 15 fps$ videos, despite the token sequence length doubling, we achieve a \textbf{10.1$\times$} acceleration in $720p, 30 fps$ generation.

\paragraph{Refinement with Condition Frames} In addition to the \textit{Text-to-Video} task, we also support the refinement for the \textit{Image-to-Video} and \textit{Video-Continuation} tasks. In the conditional coarse-to-fine generation, we first use low-resolution condition frames to generate a low-resolution video. This process can be represented as follows:
 \begin{equation}
 X_{lr} = BaseModel([Encode(X_{lr}^{cond}),\epsilon]), \epsilon \sim \mathcal{N}(\mathbf{0}, \mathbf{I}),
 \end{equation}
\begin{equation}
 X_{lr}^{cond} = Downsample(X_{hr}^{cond}),
 \end{equation}
where $X_{hr}^{cond}$ represents the high-resolution condition RGB frames, 
$X_{lr}^{cond}$ is the low-resolution condition RGB frames obtained using the spatial-temporal downsampling operation $Downsample$, and $X_{lr}$ represents the non-condition part of the low-resolution video generated in the first stage. The generation process of the refinement stage can be represented as follows:
\begin{equation}
 X_{up} = [X_{hr}^{cond}, Upsample(X_{lr})],
\end{equation}
\begin{equation}
 X_{sr} = Refinement(AddNoise(Encode(X_{up}))).
\end{equation}
At the beginning of the refinement stage, we concatenate the high-resolution version of the condition RGB frames with trilinear upsampled $X_{lr}$, this concatenation is denoted as $X_{up}$. Then, we add noise at level $t_{thresh}$ to VAE-encoded $X_{up}$. At this point, we have constructed the input for the refinement expert. The high-resolution video obtained after multiple steps of denoising is represented as $X_{sr}$. Through this design, we simultaneously support multiple tasks in refinement training, providing the coarse-to-fine generation with more application scenarios.


\subsubsection{Block Sparse Attention}
\label{sec:bsa}

\begin{figure}[htbp]
  \centering
  \includegraphics[width=1\textwidth]{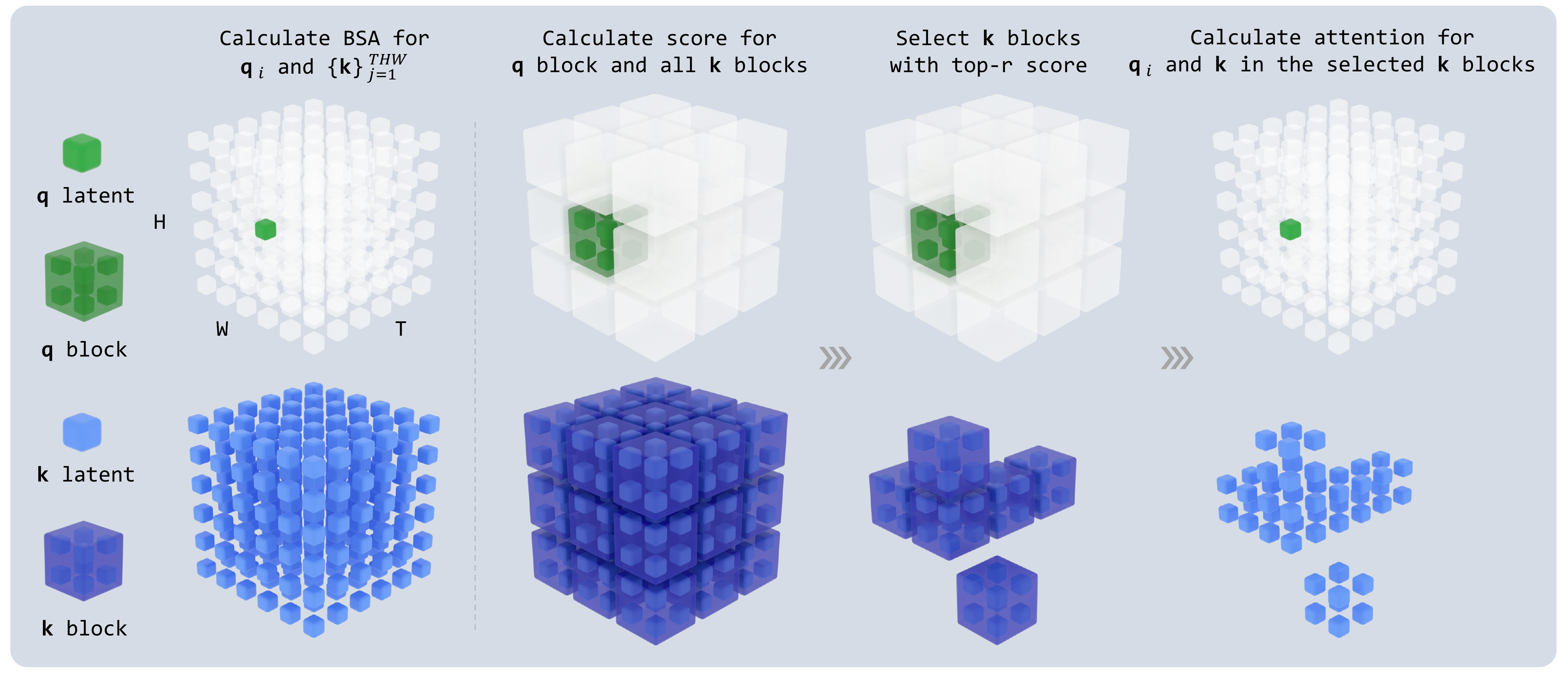}
  \caption{
    Illustration of 3D block sparse attention for query \( q_i \) and keys \( \{k_j\}_{j=1}^{T H W} \).
    \textbf{(a)} Partition \( q_i \) and all \( k_j \) into non-overlapping 3D blocks of size \( t \times h \times w \). The block containing \( q_i \) is identified, and a similarity score is computed between this query block and each key block using their average values.
    \textbf{(b)} Select the top-\( r \) key blocks with the highest similarity scores.
    \textbf{(c)} Compute the standard attention between \( q_i \) and all keys within the selected \( r \) key blocks.
    }
\end{figure}

The computational speed of both training and inference for high-resolution video generation poses a major bottleneck for practical applications, primarily due to the quadratic complexity growth of self-attention with increasing token count. 
Trainable sparse attention mechanisms have demonstrated their effectiveness in large language models~\citep{yuan2025native,lu2025moba}, and concurrent research has also validated their efficacy in video generation tasks~\citep{zhang2025vsa}.
Given the high redundancy inherent in video latent representations, we developed a trainable sparse attention operator that significantly accelerates both training and inference. By retaining less than 10\% of the original computational load, we can achieve near-lossless generation quality. Please refer to Appendix \ref{Apdx:bsa} for details. Here we highlight some key points:

\begin{itemize}
    \item Our 3D block sparse attention is open-sourced together with the base model, including both forward and backward implementations.This makes it convenient for the community to use as a modular component in their own projects.
    \item We implemented ring block sparse attention to support context parallelism (See \ref{Apdx:bsa-cp} for details), which supports efficient training of large-scale models.
    \item Users can implement other sparse attention patterns based on our implementation, such as cumulative distribution function (CDF) based or block-wise 2D+1D, by customizing the block selection mask (See \ref{Apdx:block-selection-mask} for details).
    \item In our experiments, the top-k\footnote{Note: To avoid confusion between top-k and the abbreviation 'k' for 'key', we refer to it as top-r in other parts of the report.} block sparse attention pattern achieved lossless sparse attention adaptation after training, eliminating the need for specially designed patterns; for simplicity, LongCat-Video adopted the top-k approach.
\end{itemize}

%% file: sec/4_training.tex
\section{Training}

\begin{figure}[htbp]
    \centering
    \includegraphics[width=1.0\textwidth]{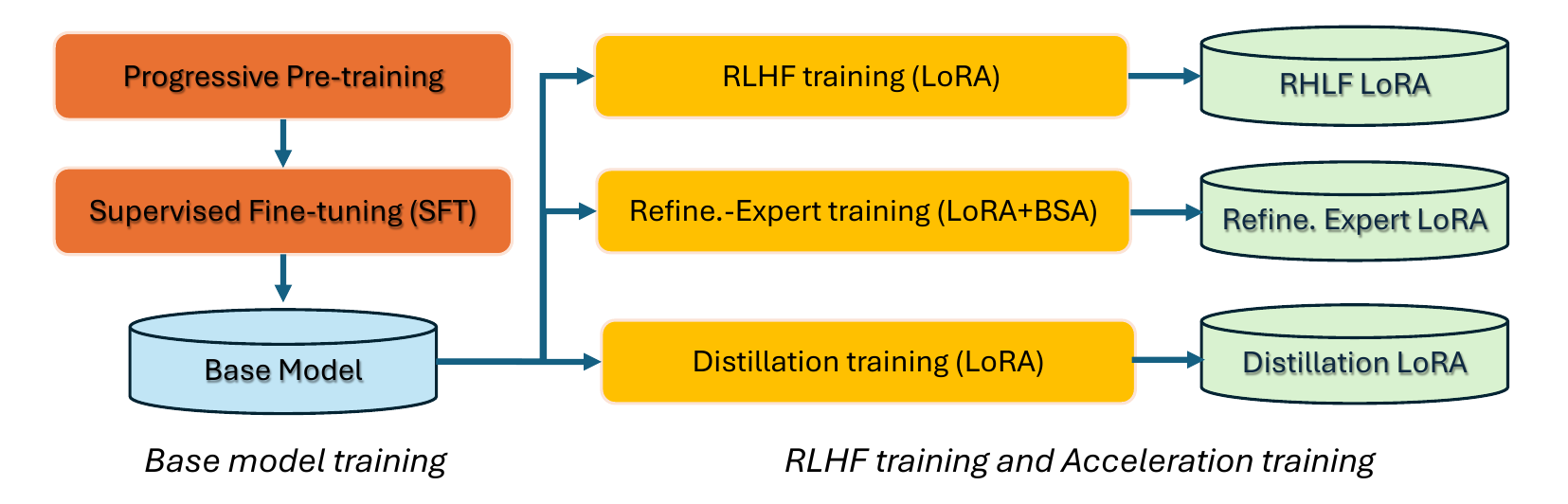}
    \caption{Overview of training process.}
    \label{fig:overview_training}
\end{figure}

As illustrated in Figure~\ref{fig:overview_training}, the overall training procedure comprises three main components. The process begins with base model training, which includes progressive pre-training and supervised fine-tuning (SFT) to produce a base video generation model. This is followed by Reinforcement Learning from Human Feedback (RLHF) training, where Group Relative Policy Optimization (GRPO) is employed to enhance model performance by aligning outputs with human preferences. The final component is acceleration training, which involves model distillation and the development of a refinement expert LoRA module for coarse-to-fine generation. For both RLHF and acceleration training, we utilize the LoRA mechanism to facilitate the stacking of various enhancements and to ensure flexibility for future extensions.

\subsection{Base Model Training}

\paragraph{Flow Matching} We employ the flow matching framework to model the diffusion process. During training, given a noise-free video latent $x_{0}$, a random noise $\epsilon \sim \mathcal{N}(\mathbf{0}, \mathbf{I})$, and a timestep $t \in [0, 1]$, the network predicts the velocity $v_{t}=\frac{dx_{t}}{dt}$ of $x_{t}$ moving towards $x_{0}$ at time $t$. $x_{t}$ can be represented as the linear interpolation as
\begin{equation}
    x_{t} = (1-t) \cdot x_{0} + t \cdot \epsilon.
\end{equation}
The ground truth velocity is
\begin{equation}
    v_{t} = x_{0} - \epsilon.
\end{equation}
The network output can be denoted as $v_{pred}(x_{t}, c, {t}; \theta)$, where $c$ represents the task conditions (text prompt, conditional image/video latents), and $\theta$ represents the model parameters. The model parameters $\theta$ are optimized by minimizing the mean squared error (MSE) between model prediction $v_{pred}$ and the ground truth velocity $v_{t}$,  denoted as a loss function
\begin{equation}
    \mathcal{L}=\mathbb{E}_{\epsilon, x_{0}, c, t}\left\|v_{pred}(x_{t}, c, {t}; \theta)-v_{t}\right\|^{2}.
\end{equation}

During training, we sample timestep $t$ from a uniform distribution, and apply a logit-normal-like loss weighting scheme. We found that this strategy is more stable than sampling timesteps directly from the logit-normal distribution. Additionally, we adaptively adjust the timestep shift based on the volume of noise tokens~\citep{esser2024scaling}, such that higher noise levels are preferred for videos with higher resolution and longer length.

\paragraph{Progressive Pre-training} During pretraining, we employ a progressive training strategy to improve efficiency, as outlined in Table~\ref{tab:progressivetraining}. The training process consists of multiple stages, beginning with model pre-training on low-resolution images to facilitate efficient learning of semantic and visual representations. After the image training stage reaches convergence, the process transitions to a dedicated video training phase, where the model captures fundamental motion dynamics. Following this, the training proceeds through several multi-task stages, during which \textit{Text-to-Image (T2I)}, \textit{Text-to-Video (T2V)}, \textit{Image-to-Video (I2V)}, and \textit{Video-Continuation (VC)} tasks are jointly optimized. For \textit{Video-Continuation (VC)} task, we also perturb conditional frames with per-frame independent noise levels~\citep{chen2024diffusionforcingnexttokenprediction} to enhance robustness to color drift. These stages progress from low-resolution to high-resolution settings. At each stage, training samples are assigned to specific size buckets according to the closest aspect ratio, thereby maximizing computational efficiency. The AdamW~\citep{loshchilov2017decoupled} optimizer is used with a constant learning rate within each stage, and the learning rate is gradually reduced as training progresses to subsequent stages.


\begin{table}[htbp]
\centering
\caption{Outline of the progressive training stages.}
\label{tab:progressivetraining}
\begin{tabular}{c|c|c|c}
\Xhline{1.5pt}
Training tasks & Size bucket & Learning rate & Iterations \\
\hline
T2I & $256p$ & 1e-4 & 285k \\
T2I + T2V & $256p \times 93$ frames & 1e-4 & 140k \\
T2I + T2V + I2V + VC & $256p \times 93$ frames & 5e-5 & 164k \\
T2I + T2V + I2V + VC & $480p \times 93$ frames & 5e-5 & 36k \\
T2I + T2V + I2V + VC & $480p + 720p \times 93$ frames & 2e-5 & 53k \\
\Xhline{1.5pt}
\end{tabular}
\end{table}

\paragraph{Supervised Fine-Tuning (SFT)} After pretraining, we conduct a supervised fine-tuning (SFT) stage using a carefully curated, high-quality dataset. The data is filtered based on multiple metrics, including aesthetic score, video quality, and motion quality, among others. To ensure balanced category representation, samples are selected inversely proportional to their density in the caption embedding space. In addition to the general high-quality dataset, we incorporate specialized datasets to further enhance the model's instruction-following capabilities, particularly for camera motion and visual style.

\begin{table}[htbp]
\centering
\caption{Specifications of supervised fine-tuning (SFT) stage.}
\label{tab:sft_stage}
\begin{tabular}{c|c|c|c}
\Xhline{1.5pt}
Training Tasks & Size Bucket & Learning rate & Iterations \\
\hline
T2I + T2V + I2V + VC & $480p + 720p \times 93$ frames & 1e-5 & 7.5k \\
\Xhline{1.5pt}
\end{tabular}
\end{table}

\subsection{RLHF Training}

After training the base model, we further improve its performance through a post-training stage that incorporates multiple video quality-related rewards using the GRPO method as described in Section~\ref{sec:method_grpo}. The key training specifications are listed in Table~\ref{tab:rhlftraining}. For the complete experimental setup, please refer to Appendix \ref{Apdx:grpo-expr}. We employ only \textit{Text-to-Video} tasks in the GRPO training, and find that the improvements of instruction-following, visual quality and motion quality generalize well to \textit{Image-to-Video} and \textit{Video-Continuation} tasks. Proposing task-specific rewards for each task (e.g. quality degradation penalty of long-video generation for \textit{Video-Continuation}) remains a future work.

\begin{table}[htbp]
\centering
\caption{Specifications of RLHF training stage.}
\label{tab:rhlftraining}
\resizebox{\textwidth}{!}{ 
\begin{tabular}{c|c|c|c|c|c|c|c}
\Xhline{1.5pt}
Training tasks & Size bucket & Group size & Prompts per step & Sampling steps & SDE steps range & Learning rate & Iterations \\ 
\hline
T2V & $480p + 720p \times 93$ frames & 4 & 64 & 16 & [0, 6] & 1e-4 & 0.5k \\
\Xhline{1.5pt}
\end{tabular}
}
\end{table}

\subsection{Acceleration Training}

 As described in Section~\ref{sec:efficient_video_gen}, we distill the model and train a refinement expert module to enable efficient inference.

\textbf{Distillation training} We have adopted Classifier-Free Guidance (CFG) distillation and consistency model (CM) distillation~\citep{ren2024hyper,wang2024phased} to enhance model inference speed. In the CFG distillation step, we distill a general negative prompt using CFG-Zero~\citep{fan2025cfg} with a default guidance strength of 4.0. The combination of CFG distillation and CM distillation enables inference with 16 steps with quality comparable to inference results with more than 50 steps. We use a LoRA training strategy to allow flexible stacking of various model enhancement and further extensions.

\begin{table}[htbp]
\centering
\caption{Specifications of distillation training.}
\label{tab:distilltraining}
\begin{tabular}{c|c|c|c|c}
\Xhline{1.5pt}
Stage & Training Tasks & Size Bucket & Learning rate & Iterations  \\
\hline
CFG distillation & T2I + T2V + I2V + VC & $480p + 720p \times 93$ frames & 5e-5 & 2k \\
CM distillation & T2I + T2V + I2V + VC & $480p + 720p \times 93$ frames & 5e-5 & 3k \\
\Xhline{1.5pt}
\end{tabular}
\end{table}

\textbf{Refinement expert training} During the refinement LoRA training process, we initially use full attention for training. Once the loss converges and stabilizes, we activate BSA to continue training. We set the sparsity of BSA to 93.75\% and the initial noise intensity for the refinement stage to 0.5. In terms of training data, we use Gray-Level Co-occurrence Matrix(GLCM)~\citep{haralick2007textural} filter to keep only data with rich texture details for training. We apply a series of degradation operations to the training data to enhance the model's ability to refine details and improve robustness. Note that we train the refinement expert on data with mixed frame rates, enabling it to support both spatial-only refinement and spatial-temporal refinement.

\begin{table}[htbp]
\centering
\caption{Specifications of refinement expert training.}
\label{tab:SRtraining}
\begin{tabular}{c|c|c|c|c|c}
\Xhline{1.5pt}
Training Stage & Sparsity & $t_{thresh}$ & Size bucket & Learning rate & Iterations  \\
\hline
 Full Attention & - & 0.5 & $720p \times 93~or~189$ frames & 5e-5 & 500 \\
 Sparse Attention & 93.75\% & 0.5 & $720p \times 93~or~189$ frames & 5e-5 & 500 \\
\Xhline{1.5pt}
\end{tabular}
\end{table}

\subsection{Training Infrastructure}

Our distributed training infrastructure incorporates mechanisms such as \textbf{DeepSpeed-Zero2}~\citep{rasley2020deepspeed}, \textbf{Context Parallelism}, \textbf{Ring Attention}, and \textbf{Activation Checkpointing}, enabling efficient training of video generation models at the 13B-parameter scale. To support mixed-resolution training, we adopt a bucket-based strategy that groups data with similar resolutions into the same bucket for batch processing. Furthermore, we employ a cache mechanism to eliminate computation bubbles arising from VAE operations across different ranks, thereby improving computational efficiency and resource utilization. These methods collectively enable the training process to achieve \textit{Model Flops Utilization} (MFU) rates ranging from 33\% to 38\%.

%% file: sec/5_evaluation.tex
\section{Evaluation}

This section presents a comprehensive evaluation of LongCat-Video's performance across multiple dimensions of video generation quality. We establish rigorous assessment protocols through both internal benchmarks and public evaluation frameworks, providing a holistic view of the model's capabilities in \textit{Text-to-Video} and \textit{Image-to-Video} generation tasks. The subsequent subsections present representative examples of LongCat-Video outputs across various video generation tasks.

\subsection{Internal Benchmarks}

We introduce an internal benchmarking suite to assess model performance across two core tasks: \textit{Text-to-Video} and \textit{Image-to-Video}. The benchmark encompasses a total of 1,628 samples, categorized into 1,228 \textit{Text-to-Video} cases~(evaluated via 500 human and 728 automatic assessments) and 400 \textit{Image-to-Video} cases. For \textit{Text-to-Video}, evaluation is conducted based on the following four key dimensions:

\begin{itemize}

  \item \textbf{Text-Alignment} evaluates whether the video comprehensively encompasses the information conveyed in the text and accurately interprets the relevant semantic expressions. It includes precise understanding of descriptions related to objects, people, scenes, styles, and other key elements.

  \item \textbf{Visual quality} is assessed from two perspectives: plausibility and realism. Plausibility focuses on the visual presentation of the video, examining whether it adheres to objective physical principles and identifying any issues such as distortion or unnatural appearances. Realism evaluates whether the scenes and subjects depicted in the video possess a sense of authenticity, aiming to avoid the presence of unrealistic elements. 

  \item \textbf{Motion quality} assesses the normalcy of motion within the video. It examines whether motion trajectories are coherent and actions are smooth, in accordance with physical laws. For human motion, object motion, and camera motion, the evaluation determines whether each type of movement reflects realistic behavior, avoiding issues such as prolonged stillness or excessive jitter.

  \item \textbf{Overall quality} represents a comprehensive quality score for the generated video based on the aforementioned sub-dimensions.

\end{itemize}
For \textit{Image-to-Video}, we further incorporate an ``Image-Alignment'' dimension in addition to the above four dimensions for evaluation:
\begin{itemize}
  \item \textbf{Image-Alignment} evaluates the extent to which the generated video faithfully preserves key attributes and relationships of both the subject and background from the reference image, while maintaining the overall style of the original reference.
\end{itemize}

\paragraph{Evaluation Protocol} The evaluation of video result in this report comprises both human and automatic model-based assessments. For human evaluation, following prior practice~\citep{gao2025seedance}, we employ two complementary methodologies: absolute Mean Opinion Score~(MOS) ratings and relative Good-Same-Bad~(GSB) assessments. The former utilizes a 5-point scale for pointwise evaluation to quantitatively measure perceptual quality across various dimensions. Detailed descriptors were established for each scoring tier to ensure metric interpretability. The final score for each model is calculated as a weighted~(2:1) average of human evaluation and automatic evaluation. The latter adopts a pairwise comparative approach, which provides more discriminative model performance rankings.

\paragraph{Quality Control} To ensure annotation quality, a comprehensive and rigorous pre-annotation training process was implemented for all annotators. Each video was independently annotated by three annotators. In cases where significant discrepancies were identified between any two annotations, two additional annotators were introduced to reassess the video. The final score for each video was derived by averaging the ratings provided by all involved annotators. This consensus-based approach enhances the reliability and objectivity of the annotation outcomes.

For automatic evaluation, we have specifically trained a vision-language judge model based on high-quality human-annotated data, capable of quantitatively evaluating text alignment, visual quality, and motion quality. Internal evaluations demonstrate that this judge model achieves correlations consistently exceeding 0.92 with human assessments across all dimensions.

\paragraph{Data Taxonomy for Text-to-Video Evaluation}
Our text-to-video evaluation benchmark comprises two distinct subsets: 500 prompts designed for human evaluation and 728 for automatic evaluation. The human evaluation subset is characterized by its exceptional semantic diversity, spanning 48 distinct categories. This design ensures a balanced assessment, preventing the overrepresentation of any single capability, with the most frequent category constituting only 39.2\% of the prompts. Critically, the benchmark features a long tail of specialized tasks: 58.3\% of categories appear with a frequency of 5\% or less. These range from foundational abilities such as \textit{Entity Generation} and \textit{Action} to complex functions like \textit{Physical Simulation} and \textit{Inductive Reasoning}. Furthermore, the prompts exhibit significant structural diversity. Their lengths follow a pronounced bimodal distribution: 34.8\% are concise ($\leq$20 words) and 34.6\% are highly detailed ($\geq$51 words), with an overall range of 4 to 121 words. To ensure comprehensive coverage for the automatic evaluation subset, we curate prompts from high-quality public datasets, including T2VCompbench~\citep{sun2025t2v} and MovieGen~\citep{polyak2024movie}, and supplement them with in-house prompts to cover a wide array of video generation scenarios.

\paragraph{Text-to-Video Evaluation} Leveraging our internal benchmark, we first conducted a comprehensive comparative evaluation of LongCat-Video against several leading video generation models in text-to-video setting. Specifically, we compare with two advanced proprietary models Veo3~\citep{veo} and PixVerse-V5~\citep{PixVerse}, as well as the current SOTA open-source model Wan 2.2-T2V-A14B~\citep{wan2025wan}.

\begin{figure}[htbp]
  \centering
  \includegraphics[width=1.0\textwidth]{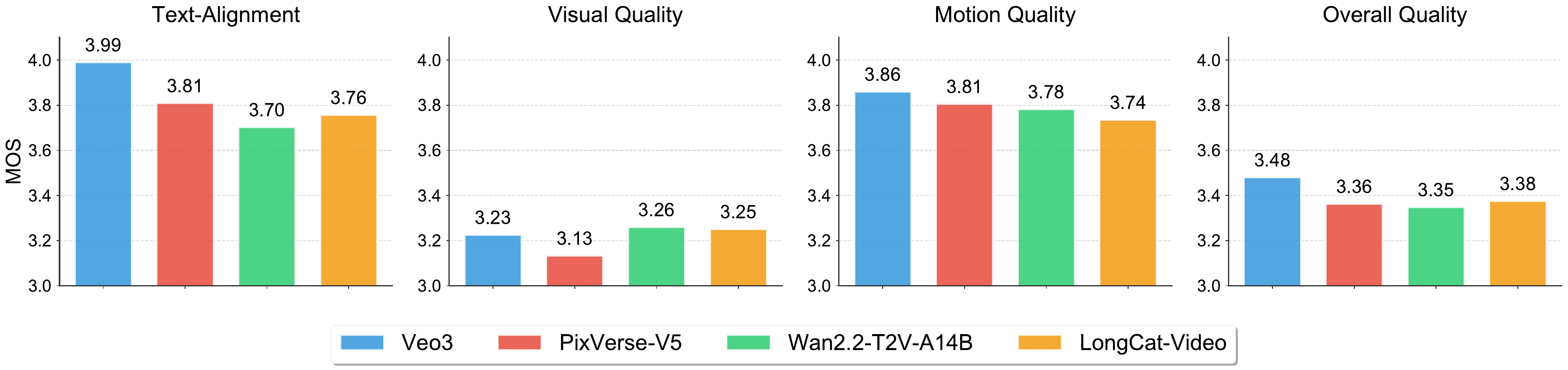}
  \caption{Text-to-Video MOS evaluation results on our internal benchmark.}
  \label{fig:t2v_results}
\end{figure}

The MOS evaluation results are illustrated in Figure~\ref{fig:t2v_results}. Our analysis reveals that LongCat-Video demonstrates a highly competitive and well-balanced performance. A standout achievement is its excellence in Visual Quality, where it achieves a score that is nearly on par with the top performer, Wan 2.2, and significantly surpasses PixVerse-V5, which shows a clear deficit in this area.
In terms of Overall Quality, LongCat-Video establishes itself as a top-tier model, achieving a score superior to both PixVerse-V5 and Wan 2.2-T2V-A14B. While Veo3 leads in this category, its advantage is built upon superior text-alignment and motion scores. In contrast, our model provides a more consistent, high-quality experience. For Text-Alignment, LongCat-Video delivers robust results, proving its strong capability in semantic understanding, though Veo3 sets a particularly high benchmark.

\begin{figure}[htbp]
  \centering
  \includegraphics[width=0.98\textwidth]{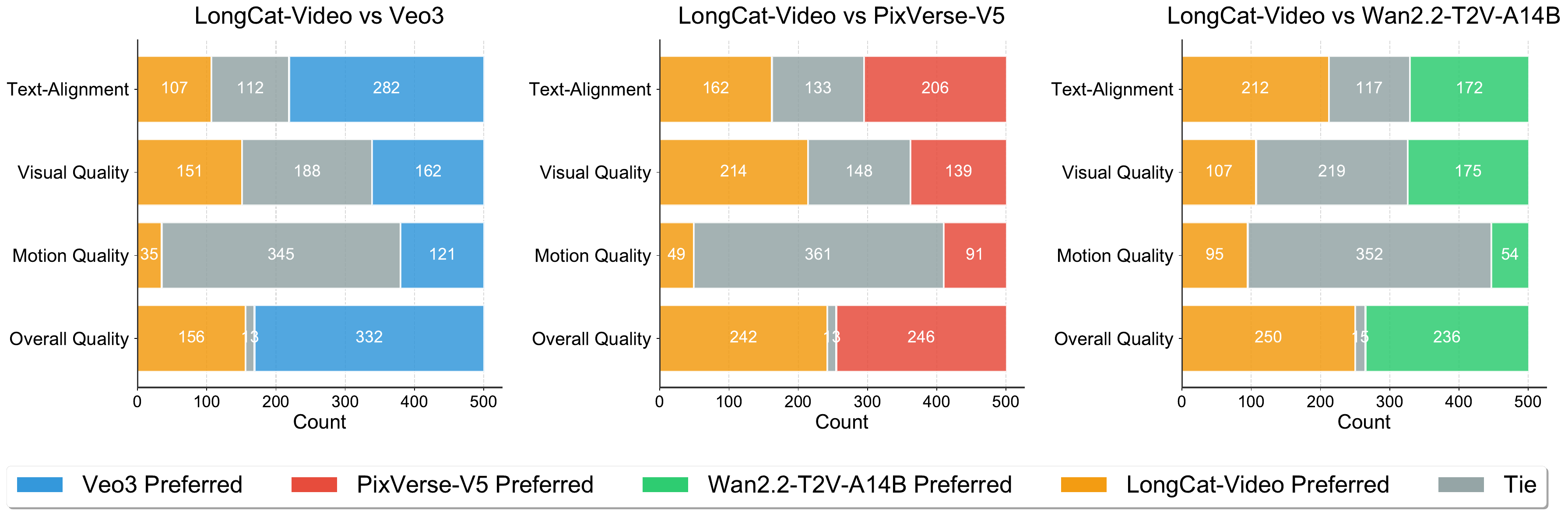}
  \caption{Text-to-Video GSB evaluation results on our internal benchmark.}
  \label{fig:t2v_gsb}
\end{figure}

The GSB evaluation results are shown in Figure~\ref{fig:t2v_gsb}. The user preference study indicates that LongCat-Video's performance, while trailing the state-of-the-art closed-source model Veo3, is highly competitive and on par with other leading proprietary models like PixVerse-V5. In the direct comparison, LongCat-Video and PixVerse-V5 are nearly tied in overall quality~(242 vs. 246), with our model demonstrating a distinct advantage in visual quality. More importantly, when benchmarked against the current state-of-the-art open-source model, Wan2.2-T2V-A14B, our model shows a clear superiority. LongCat-Video was preferred by users in overall quality, driven by significant leads in both text-alignment and motion quality.


\paragraph{Data Taxonomy for Image-to-Video Evaluation}
Our benchmark for \textit{Image-to-Video} evaluation is built upon a curated set of 100 first-frame reference images, designed to exhibit comprehensive diversity across multiple dimensions. These dimensions include \textbf{style} (e.g., photorealism, ink wash, 2D/3D animation, oil painting, sketch), \textbf{content} (e.g., human subjects, animals, plants, food, vehicles, indoor/outdoor environments), and \textbf{quality} (high vs. standard). Each image is further defined by metadata such as aspect ratios (1:1, 16:9, 9:16) and resolutions ($720p$, $1080p$, $2K$). To rigorously evaluate model sensitivity and dependency, each reference image is paired with a set of four distinct prompt types: (1) \textit{detailed prompts} that specify fine-grained attributes; (2) \textit{concise prompts} with minimal instructions; (3) \textit{contradictory prompts} designed to conflict with the visual reference; and (4) \textit{empty prompts} to assess unconditional generation based on the image. This quadripartite prompt structure enables a robust assessment of the model's cross-modal alignment and generative capabilities.

\paragraph{Image-to-Video Evaluation} We then compare LongCat-Video against several leading video generation models in image-to-video generation setting. Concretely, we compare with two advanced proprietary models Seedance 1.0~\citep{gao2025seedance} and Hailuo-2, as well as the current SOTA open-source model Wan 2.2-I2V-A14B~\citep{wan2025wan}. 

\begin{figure}[htbp]
  \centering
  \includegraphics[width=1.0\textwidth]{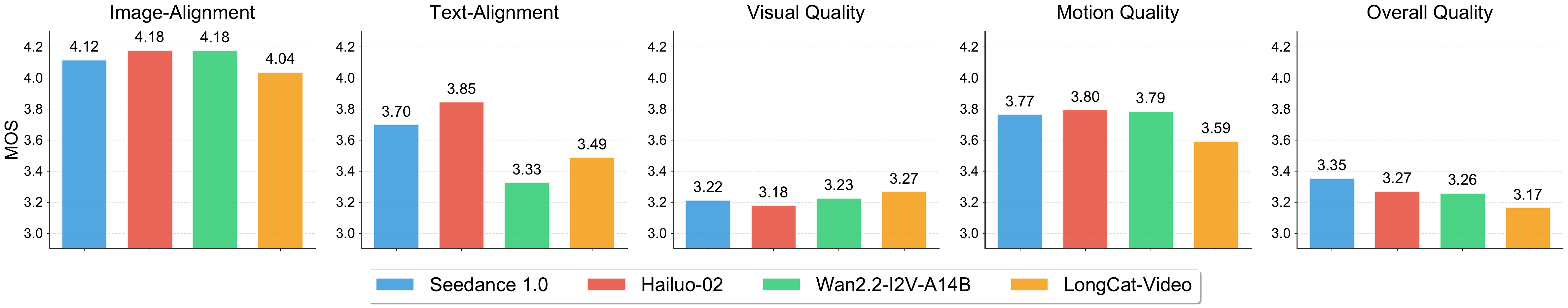}
  \caption{Image-to-Video MOS evaluation results on our internal benchmark.}
  \label{fig:i2v_results}
\end{figure}

The MOS evaluation results are illustrated in Figure~\ref{fig:i2v_results}. As shown in the figure, LongCat-Video achieves the highest score in Visual Quality~(3.27), indicating its strength in generating aesthetically pleasing frames. However, it scores lower on Image-Alignment~(4.04) and Motion Quality~(3.59) compared to the other models. Hailuo-02 and Wan2.2-I2V-A14B perform best in Image-Alignment~(4.18), while Hailuo-02 leads in Motion Quality~(3.80). In the Overall Quality evaluation, LongCat-Video~(3.17) is rated as competitive, though it trails the other models, with Seedance 1.0 achieving the highest overall score of 3.35. This suggests that while our model excels in visual fidelity, there is room for improvement in maintaining temporal consistency and alignment with the source image.

\subsection{Public Benchmarks}

As a supplement to internal benchmarks, we also evaluated LongCat-Video on the widely used public benchmark VBench~\citep{huang2024vbench,zheng2025vbench}. Specifically, we conducted assessments on the latest version of VBench 2.0. The evaluation results are shown Table~\ref{tab:t2v_vbench_2_0_benchmark}. On VBench 2.0, Long-Cat Video also demonstrated strong performance, with a total score second only to Veo3~\citep{veo} and Vidu Q1~\citep{Vidu}. It is noteworthy that LongCat-Video led all other methods in the \textit{Commonsense} dimension, indicating that our approach excels in aspects such as motion rationality and physical laws. This aligns with Long-Cat Video's outstanding long video generation capabilities and represents a key advantage in moving towards world model development.

\begin{table}[htbp]
\centering
\caption{Text-to-Video evaluation results on VBench 2.0 benchmark.}
\label{tab:t2v_vbench_2_0_benchmark}
\resizebox{\textwidth}{!}{ 
\begin{tabular}{c|c|c|c|c|c|c|c|c}
\Xhline{1.5pt}
Model name & Accessibility & Evaluation Date & Creativity$\uparrow$ & Commonsense$\uparrow$ & Controllability$\uparrow$ & Human Fidelity$\uparrow$ & Physics$\uparrow$ & Total Score$\uparrow$ \\
\hline
HunyuanVideo~\citep{kong2024hunyuanvideo} & Open Source & 2025-03 & 41.84\% & 63.44\% & 28.60\% & 82.41\% & 60.20\% & 55.30\%  \\
Wan2.1~\citep{wan2025wan} & Open Source & 2025-03 & 55.25\% & 63.98\% & 37.32\% & 81.60\% & 62.84\% & 60.20\%  \\
Sora-480p~\citep{sora} & Proprietary & 2025-03 & 60.57\% & 64.32\% & 22.09\% & 87.72\% & 57.18\% & 58.38\%  \\
Kling1.6~\citep{Kling} & Proprietary & 2025-03 & 48.58\% & 65.45\% & 33.05\% & 83.56\% & 64.35\% & 59.00\%  \\
Vidu Q1~\citep{Vidu} & Proprietary & 2025-04 & 56.54\% & 65.98\% & 38.13\% & 81.24\% & 71.63\% & 62.70\%  \\
Seedance 1.0 Pro~\citep{gao2025seedance} & Proprietary & 2025-06 & 53.04\% & 64.31\% & 39.84\% & 77.06\% & 64.81\% & 59.81\%  \\
Veo3~\citep{veo} & Proprietary & 2025-09 & 60.85\% & 69.48\% & 47.04\% & 86.88\% & 69.35\% & 66.72\%  \\
\hline
LongCat-Video & Open Source & 2025-10 & 54.73\% & 70.94\% & 44.79\% & 80.20\% & 59.92\% & 62.11\%  \\
\Xhline{1.5pt}
\end{tabular}
}
\end{table}

\clearpage

\subsection{Text-to-Video Examples}

\begin{figure}[htbp]
  \centering
  \includegraphics[width=0.99\textwidth]{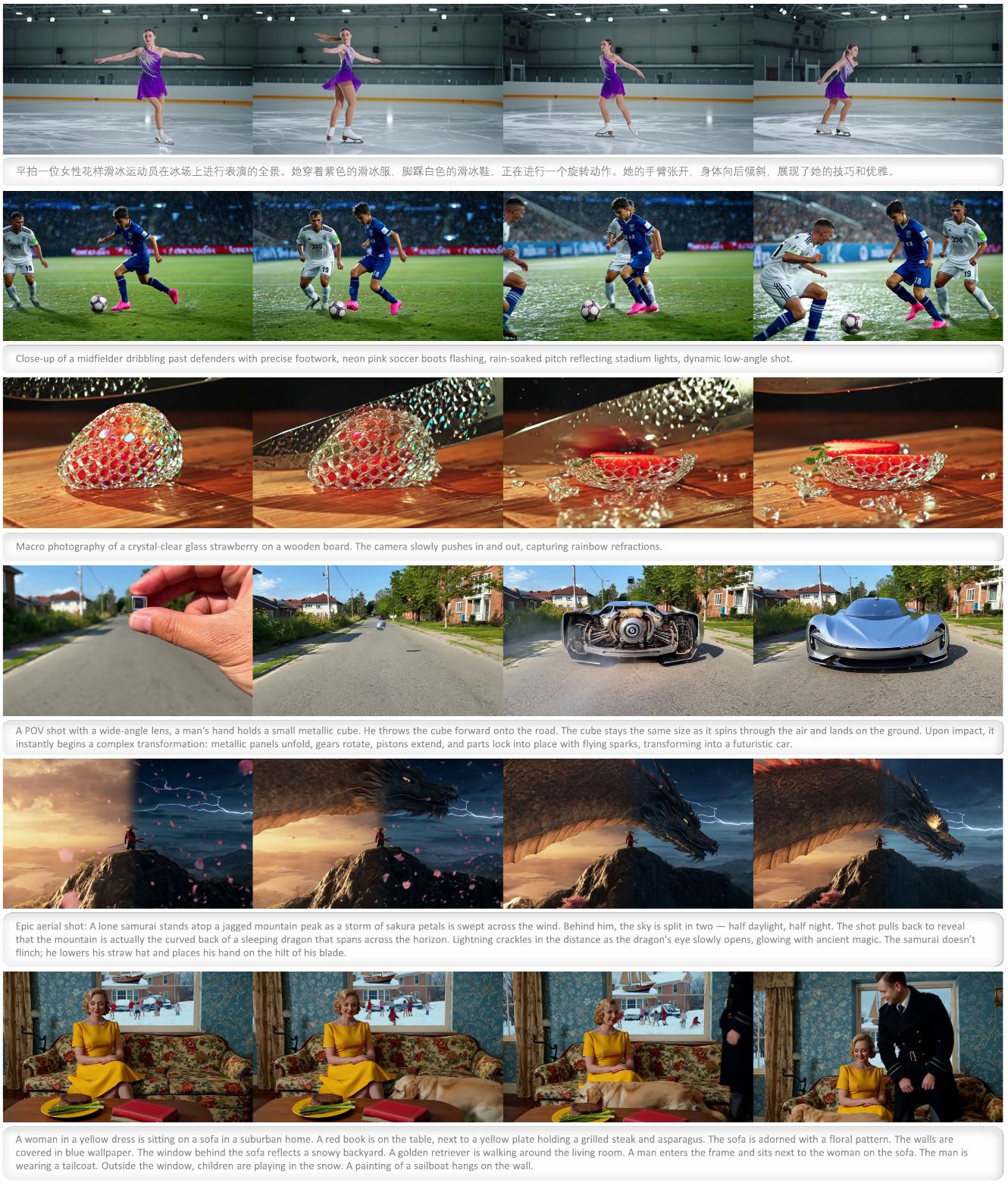}
  \caption{Results on \textit{Text-to-Video} generation.}
\end{figure}

\clearpage

\subsection{Image-to-Video Examples}
\begin{figure}[H]
  \centering
  \includegraphics[width=0.99\textwidth]{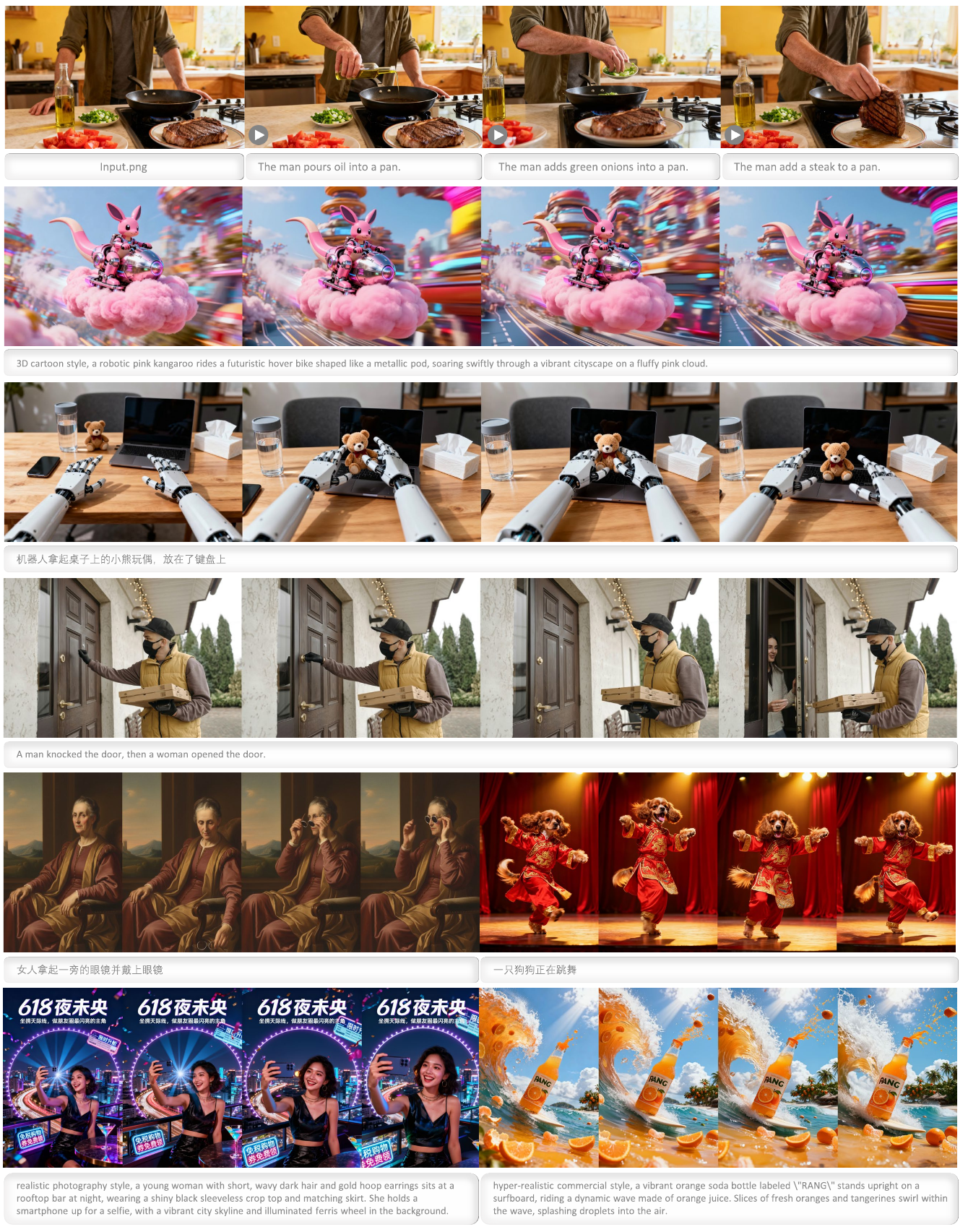}
  \caption{Results on \textit{Image-to-Video}. As shown in the top row, given the same initial image, LongCat-Video accurately responds to instructions for various actions.}
\end{figure}

\clearpage

\subsection{Long-Video Generation Examples}

\begin{figure}[H]
  \centering
  \includegraphics[width=\textwidth]{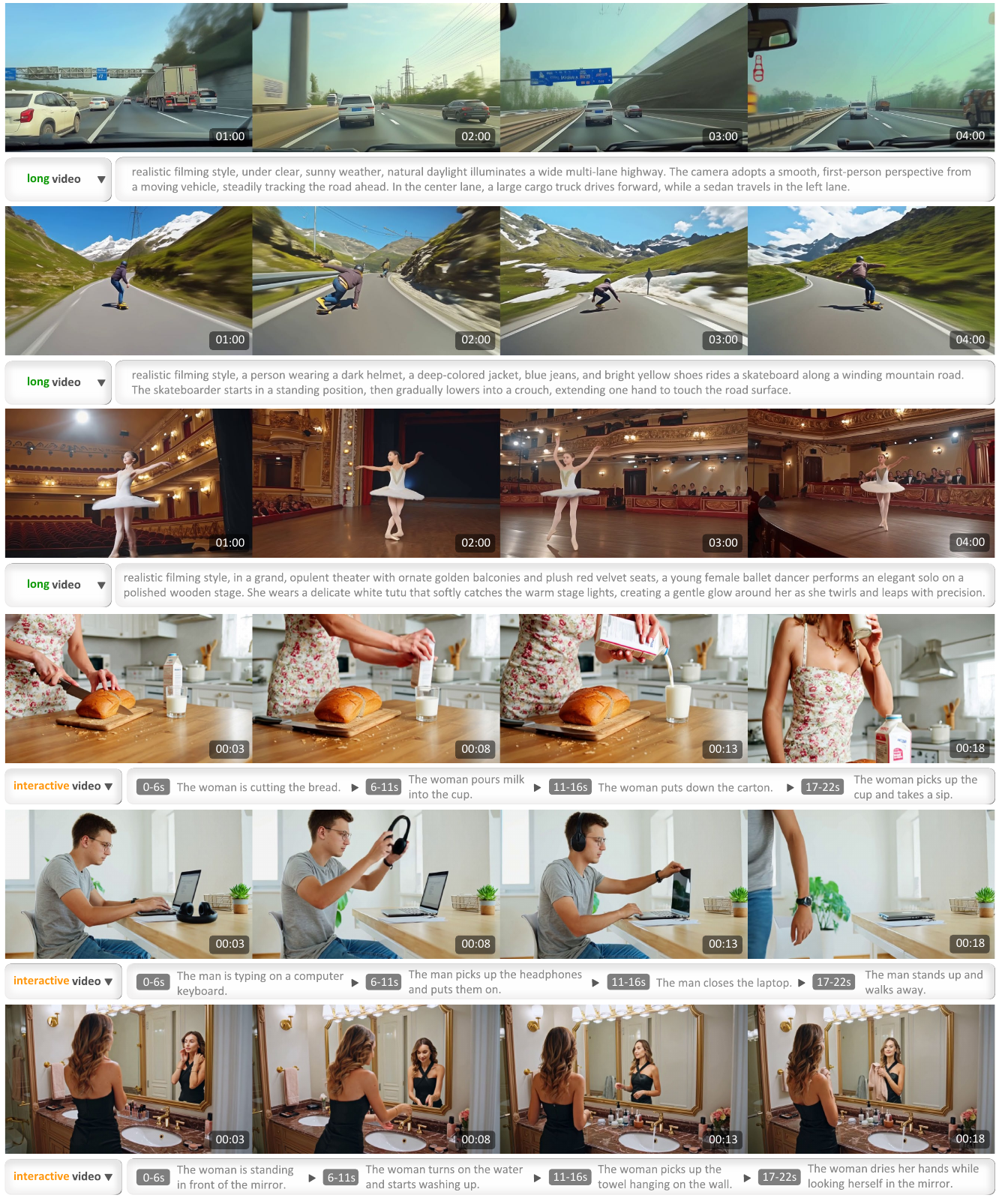}
  \caption{Results on \textit{Video-Continuation}. LongCat-Video supports minutes-long video generation without quality degradation, as well as interactive video generation with changing instructions for each clip.}
\end{figure}

\clearpage

%% file: sec/6_conclusion.tex
\section{Conclusion and Future Work}

We introduce LongCat-Video, a 13B-parameter foundational video generation model that unifies \textit{Text-to-Video}, \textit{Image-to-Video}, and \textit{Video-Continuation} tasks within a single framework. LongCat-Video demonstrates strong performance across all supported tasks, particularly excelling in long video generation, which is enabled by pretraining on the \textit{Video-Continuation} task. As a robust general-purpose video generation model, LongCat-Video is applicable to a wide range of video content creation scenarios. Moreover, it marks our first step toward developing world models. Efficient long video generation addresses the rendering problem of world models, enabling models to express their world knowledge through generated video content. Future directions include better modeling of physical knowledge, multi-modal memory integration in video generation, and the incorporation of knowledge from LLM and MLLM.

%% file: sec/7_contributors.tex
\section{Contributors and Acknowledgments}

Contributors are listed in alphabetical order by their last names.  Names marked with an asterisk (*) indicate people who have left our
team.

\paragraph{Contributors}\mbox{}\\[1\baselineskip]
\begin{tabular}{p{0.19\textwidth}p{0.19\textwidth}p{0.19\textwidth}p{0.19\textwidth}p{0.19\textwidth}}
Xunliang Cai & Qilong Huang & Zhuoliang Kang & Hongyu Li & Shijun Liang \\
Liya Ma & Siyu Ren & Xiaoming Wei & Rixu Xie & Tong Zhang \\
\end{tabular}

\paragraph{Acknowledgments}\mbox{}\\[1\baselineskip]
\begin{tabular}{p{0.19\textwidth}p{0.19\textwidth}p{0.19\textwidth}p{0.19\textwidth}p{0.19\textwidth}}
Xuezhi Cao & Hui Chen & Fengjiao Chen & Tianye Dai & Feng Gao \\
Ying Guo* & Xiaoyu Li & Shengxi Li & Hao Lu & Xiaofeng Mei* \\
Zhuqi Mi & Xin Pan & Liang Shi & Yuchen Tang & Chao Wang \\
Ziwen Wang  & Wei Yi & Yong Zhang & Zizhe Zhao \\
\end{tabular}

%% file: sec/8_appendix.tex
\section{Appendix}\label{appendix}

\subsection{Appendix-A}

\subsubsection{GRPO Preliminaries}
\label{Apdx:grpo-prelim}
The GRPO method optimizes the generative flow model by maximizing the following objective function:

\begin{equation}
\mathcal{J}_{\mathrm{GRPO}}(\theta)=\mathbb{E}_{c \sim \mathcal{C},\left\{x^i\right\}_{i=1}^G \sim \pi_{\theta_{\text{old }}}(\cdot \mid c)}\left[\frac{1}{G} \sum_{i=1}^G \frac{1}{T} \sum_{t=0}^{T-1}\left(\mathcal{L}_{\text{policy }}(\theta)-\beta D_{\mathrm{KL}}\left(\pi_\theta \| \pi_{\text{ref }}\right)\right)\right],
\label{eq:grpo_objective}
\end{equation}

Below we elaborate on each component of this objective.

\paragraph{Sampling Process.} A group of $G$ samples $\left\{\boldsymbol{x}^i\right\}_{i=1}^G$ is drawn from the current policy $\pi_{\theta_{\text{old }}}$ conditioned on the prompt $c$. Each sample is generated by discretizing the reverse-time stochastic differential equation (SDE):

\begin{equation}
x_{t+\Delta t}=x_t+\left[v_\theta\left(x_t, t, c\right)+\frac{\sigma_t^2}{2 t}\left(x_t+(1-t) v_\theta\left(x_t, t, c\right)\right)\right] \Delta t+\sigma_t \sqrt{\Delta t} \epsilon,
\label{eq:sde_discretization}
\end{equation}

with $\epsilon \sim \mathcal{N}(0, \mathbf{I})$ and noise schedule $\sigma_t=a \sqrt{t /(1-t)}$. This process yields complete trajectories $\left\{\left(x_T^i, x_{T-1}^i, \cdots, x_0^i\right)\right\}_{i=1}^G$ for policy optimization.

\paragraph{Policy Loss.} The policy loss $\mathcal{L}_{\text{policy }}(\theta)=r_t^i(\theta) \hat{A}_t^i$ consists of two elements:

1) Importance ratio: $r_t^i(\theta)=\frac{p_\theta\left(x_{t-1}^i \mid x_t^i, c\right)}{p_{\theta_{\text{old }}}\left(x_{t-1}^i \mid x_t^i, c\right)}$ quantifies the probability change for transition $x_t^i \rightarrow x_{t-1}^i$ between policy updates, where the transition probability follows:
\begin{equation}
p_\theta\left(x_{t-1} \mid x_t, c\right)=\mathcal{N}\left(x_{t-1}; \mu_\theta\left(x_t, t, c\right), \sigma_t^2 \Delta t \mathbf{I}\right).
\label{eq:transition_prob}
\end{equation}

2) Group-relative advantage: $\hat{A}_t^i=\frac{R\left(x_0^i, c\right)-\operatorname{mean}\left(\left\{R\left(x_0^j, c\right)\right\}_{j=1}^G\right)}{\operatorname{std}\left(\left\{R\left(x_0^j, c\right)\right\}_{j=1}^G\right)}$ provides normalized advantage estimates by comparing individual rewards against group statistics.

\paragraph{KL Regularization.} The KL divergence term $D_{\mathrm{KL}}\left(\pi_\theta \| \pi_{\mathrm{ref}}\right)$ ensures training stability by constraining policy deviation from the reference policy. For the flow matching formulation, this term can be expressed as:

\begin{equation}
D_{\mathrm{KL}}\left(\pi_\theta \| \pi_{\mathrm{ref}}\right)=\frac{\Delta t}{2}\left(\frac{\sigma_t(1-t)}{2 t}+\frac{1}{\sigma_t}\right)^2\left\|v_\theta\left(x_t, t, c\right)-v_{\mathrm{ref}}\left(x_t, t, c\right)\right\|^2,
\label{eq:kl_analytical}
\end{equation}
with $\beta$ controlling the regularization strength.

\subsubsection{The Gradient of the Policy and KL Loss}
\label{Apdx:grpo-rew}

We derive the gradient of the policy loss $\mathcal{L}_{\text{policy}}(\theta) = r_t^i(\theta) \hat{A}_t^i$ with respect to the parameters $\theta$, The gradient computation proceeds as follows:
\[
\nabla_\theta \mathcal{L}_{\text{policy}}(\theta) = \hat{A}_t^i \nabla_\theta r_t^i(\theta).
\]

\[
\nabla_\theta r_t^i(\theta) = \frac{p_\theta\left(x_{t-1}^i \mid x_t^i, c\right)}{p_{\theta_{\text{old}}}\left(x_{t-1}^i \mid x_t^i, c\right)} \nabla_\theta \log p_\theta\left(x_{t-1}^i \mid x_t^i, c\right)=\nabla_\theta \log p_\theta\left(x_{t-1}^i \mid x_t^i, c\right).
\]

Combining these results gives the policy gradient:
\begin{equation}
\nabla_\theta \mathcal{L}_{\text{policy}}(\theta) = \hat{A}_t^i r_t^i(\theta) \nabla_\theta \log p_\theta\left(x_{t-1}^i \mid x_t^i, c\right).
\end{equation}

We now compute the score function $\nabla_\theta \log p_\theta\left(x_{t-1} \mid x_t, c\right)$. The conditional distribution is Gaussian:
\[
p_\theta\left(x_{t-1} \mid x_t, c\right) = \mathcal{N}\left(x_{t-1}; \mu_\theta\left(x_t, t, c\right), \sigma_t^2 \Delta t I\right).
\]

\[
\nabla_\theta \log p_\theta = \frac{1}{\sigma_t^2 \Delta t} \left(x_{t-1} - \mu_\theta\right) \cdot \nabla_\theta \mu_\theta.
\]

From the SDE sampling process, we have the reparameterization:
\[
x_{t-1} = \mu_\theta + \sigma_t \sqrt{\Delta t} \epsilon, \quad \epsilon \sim \mathcal{N}(0, I),
\]
Substituting:
\[
\nabla_\theta \log p_\theta = \frac{1}{\sigma_t^2 \Delta t} \left(\sigma_t \sqrt{\Delta t} \epsilon\right) \cdot \nabla_\theta \mu_\theta = \frac{1}{\sigma_t \sqrt{\Delta t}} \epsilon \cdot \nabla_\theta \mu_\theta.
\]

\begin{equation}
\mu_\theta = x_t + \left[v_\theta\left(x_t, t, c\right) + \frac{\sigma_t^2}{2t}\left(x_t + (1-t) v_\theta\left(x_t, t, c\right)\right)\right](-\Delta t)
\end{equation}

Simplifying the drift term:

\begin{equation}
\begin{split}
\mathrm{drift} &= v_\theta + \frac{\sigma_t^2}{2t} x_t + \frac{\sigma_t^2}{2t}(1-t) v_\theta \\
&= v_\theta\left(1 + \frac{\sigma_t^2(1-t)}{2t}\right) + \frac{\sigma_t^2}{2t} x_t
\end{split}
\end{equation}

Thus:

\begin{equation}
\mu_\theta = x_t - \Delta t \cdot \mathrm{drift}
\end{equation}

Taking the gradient with respect to $\theta$ (noting that $x_t$ is constant):

\begin{equation}
\nabla_\theta \mu_\theta = -\Delta t \cdot \nabla_\theta \mathrm{drift} = -\Delta t \cdot \left(1 + \frac{\sigma_t^2(1-t)}{2t}\right) \nabla_\theta v_\theta
\end{equation}

Substituting into $\nabla_\theta \log p_\theta$:

\begin{equation}
\begin{split}
\nabla_\theta \log p_\theta &= \frac{1}{\sigma_t \sqrt{\Delta t}} \epsilon \cdot \left[-\Delta t \cdot \left(1 + \frac{\sigma_t^2(1-t)}{2t}\right) \nabla_\theta v_\theta\right] \\
&= -\frac{\sqrt{\Delta t}}{\sigma_t} \left(1 + \frac{\sigma_t^2(1-t)}{2t}\right) \epsilon \cdot \nabla_\theta v_\theta
\end{split}
\end{equation}

Therefore, the gradient of the policy loss is:

\begin{equation}
\nabla_\theta \mathcal{L}_{\text{policy}}(\theta) = \hat{A}_t^i r_t^i(\theta) \cdot \left[-\frac{\sqrt{\Delta t}}{\sigma_t} \left(1 + \frac{\sigma_t^2(1-t)}{2t}\right) \epsilon \cdot \nabla_\theta v_\theta\right]
\end{equation}

Now, we substitute $a=1$ and $\sigma_t = \sqrt{\frac{t}{1-t}}$ (so $\sigma_t^2 = \frac{t}{1-t}$). Computing the coefficient term:

\begin{equation}
1 + \frac{\sigma_t^2(1-t)}{2t} = 1 + \frac{\frac{t}{1-t} \cdot (1-t)}{2t} = 1 + \frac{1}{2} = \frac{3}{2}
\end{equation}

And the scaling term:

\begin{equation}
\frac{\sqrt{\Delta t}}{\sigma_t} = \frac{\sqrt{\Delta t}}{\sqrt{\frac{t}{1-t}}} = \sqrt{\Delta t} \cdot \sqrt{\frac{1-t}{t}} = \sqrt{\frac{\Delta t(1-t)}{t}}
\end{equation}

Substituting these simplifications, we obtain the final policy gradient expression:

\begin{equation}
\nabla_\theta \mathcal{L}_{\text{policy}}(\theta) = -\frac{3}{2} \hat{A}_t^i \sqrt{\frac{\Delta t(1-t)}{t}} \epsilon \cdot \nabla_\theta v_\theta
\end{equation}

By introducing a reweighting coefficient defined as:

\begin{equation}
\lambda_{\mathrm{policy}}(t, \Delta t) = \kappa(t, \Delta t)^{-1} = \sqrt{\frac{t}{\Delta t(1-t)}}
\end{equation}

The reweighted policy loss becomes:

\begin{equation}
\mathcal{L}_{\text{policy, reweighted}}(\theta) = \lambda_{\mathrm{policy}}(t, \Delta t) \cdot \mathcal{L}_{\text{policy}}(\theta)
\end{equation}

This yields the modified gradient:

\begin{equation}
\label{eq:policy_reweight}
\nabla_\theta \mathcal{L}_{\text{policy, reweighted}}(\theta) = -\frac{3}{2} \hat{A}_t^i  \cdot \epsilon \cdot \nabla_\theta v_\theta
\end{equation}

Similarly, the gradient of the KL divergence term can be derived as:

\begin{equation}
\nabla_\theta D_{\mathrm{KL}}(\theta)  = \Delta t \cdot \frac{9}{4} \cdot \frac{1-t}{t} \cdot (v_\theta - v_{\mathrm{ref}}) \cdot \nabla_\theta v_\theta
\end{equation}

This expression reveals that the KL loss gradient suffers from the same scaling issues as the policy loss gradient. To address this, we also introduce a KL reweighting coefficient:

\begin{equation}
\lambda_{\text{KL}}(t, \Delta t) = k_{\mathrm{KL}}(t, \Delta t)^{-1} = \frac{t}{\Delta t(1-t)}
\end{equation}

The reweighted KL loss becomes:

\begin{equation}
\mathcal{L}_{\text{KL, reweighted}}(\theta)  = \lambda_{\mathrm{KL}}(t, \Delta t) \cdot D_{\mathrm{KL}}(\theta) 
\end{equation}

yielding the simplified gradient:

\begin{equation}
\nabla_\theta \mathcal{L}_{\mathrm{KL, reweighted}}(\theta)  = \frac{9}{4} \cdot (v_\theta - v_{\mathrm{ref}}) \cdot \nabla_\theta v_\theta
\end{equation}

Based on the reweighting coefficients for the policy loss and KL loss, the revised GRPO objective function is as follows:

\begin{equation}
\begin{split}
\mathcal{J}_{\mathrm{GRPO}}(\theta) = \mathbb{E}_{
    \substack{
        c \sim \mathcal{C},\ t' \sim \mathcal{U}(0, T'-1), \\ 
        \{\boldsymbol{x}^i\}_{i=1}^G \sim \pi_{\theta_{\text{old}}}(\cdot \mid c, t')
    }
}
\bigg[ \frac{1}{G} \sum_{i=1}^G \bigg( & \lambda_{\text{policy}}(\frac{t'}{T}, \Delta \frac{t'}{T}) \cdot \mathcal{L}_{\text{policy}}(\theta) - \beta \lambda_{\mathrm{KL}}(\frac{t'}{T}, \Delta \frac{t'}{T}) \cdot D_{\mathrm{KL}}\left(\pi_\theta \| \pi_{\text{ref}}\right) \bigg) \bigg]
\end{split}
\end{equation}

\subsubsection{Fix the stochastic timestep in SDE sampling}
\label{Apdx:grpo-fix-sde}
As described in Para. "Fix the stochastic timestep in SDE sampling" in Sec. \ref{sec:grpo-fixsde}, the objective function is accordingly simplified to focus only on the critical stochastic timestep:

\begin{equation}
\mathcal{J}_{\mathrm{GRPO\text{-}Selective}}(\theta) = 
\mathbb{E}_{c \sim \mathcal{C},\ t^{\prime} \sim \mathcal{U}(0, T^{\prime}-1),\ \{x^i\}_{i=1}^G \sim \pi_{\mathrm{old}}(\cdot \mid c, t^{\prime})} 
\left[\frac{1}{G} \sum_{i=1}^G \left( r_{t^{\prime}}^i(\theta) \hat{A}^i - \beta D_{\mathrm{KL}}\left( \pi_\theta \| \pi_{\mathrm{ref}} \right)_{t^{\prime}} \right) \right],
\end{equation}

\noindent where $t^{\prime} \sim \mathcal{U}(0, T^{\prime}-1)$ indicates uniform sampling of the critical timestep from the first $T^{\prime}$ steps. We set $T^{\prime}=6$ in our experiments. (The total sampling steps for training is set to 16.)

\subsubsection{Multi-reward GRPO Training}
\label{Apdx:grpo-multireward}
Eq.(\ref{eq:policy_reweight}) reveals that in flow matching models, GRPO fundamentally uses the relative advantage $\hat{A}_t^i$ and the noise term $\epsilon$ to estimate the gradient of the reward with respect to the velocity field, following the chain rule decomposition:

\begin{equation}
\frac{d R}{d \theta} = \frac{d R}{d v_\theta} \cdot \frac{d v_\theta}{d \theta}
\end{equation}

where the GRPO framework provides the specific form:

\begin{equation}
\frac{d R}{d v_\theta} \approx -\frac{3}{2} \hat{A}_t^i \cdot \epsilon
\end{equation}

When optimizing for multiple reward functions $R_1, R_2, \ldots, R_n$ with corresponding weights $w_1, w_2, \ldots, w_n$, the total gradient is given by the weighted sum:

\begin{equation}
\nabla_\theta J_{\text{total}} = \sum_{k=1}^n w_k \cdot \frac{d R_k}{d \theta}
\end{equation}

Applying the chain rule decomposition for each reward:

\begin{equation}
\nabla_\theta J_{\text{total}} = \sum_{k=1}^n w_k \cdot \left( \frac{d R_k}{d v_\theta} \cdot \frac{d v_\theta}{d \theta} \right) = \left( \sum_{k=1}^n w_k \cdot \frac{d R_k}{d v_\theta} \right) \cdot \frac{d v_\theta}{d \theta}
\end{equation}

Substituting the GRPO expression for each reward gradient:

\begin{equation}
\nabla_\theta J_{\text{total}} = \left( \sum_{k=1}^n w_k \cdot \left( -\frac{3}{2} \hat{A}_{k,t}^i \cdot \epsilon \right) \right) \cdot \frac{d v_\theta}{d \theta} = -\frac{3}{2} \left( \sum_{k=1}^n w_k \cdot \hat{A}_{k,t}^i \right) \cdot \epsilon \cdot \nabla_\theta v_\theta
\end{equation}

This demonstrates that the effective relative advantage in the policy loss for multi-reward optimization is exactly the weighted sum of the individual relative advantages. Therefore, the corresponding policy loss becomes:

\begin{equation}
\mathcal{L}_{\text{policy, multi}}(\theta) = r_t^i(\theta) \cdot \left( \sum_{k=1}^n w_k \cdot \hat{A}_{k,t}^i \right)
\end{equation}

where each relative advantage $\hat{A}_{k,t}^i$ is computed independently for reward $R_k$ using group normalization:

\begin{equation}
\hat{A}_{k,t}^i = \frac{R_k\left(\boldsymbol{x}_0^i, \boldsymbol{c}\right) - \operatorname{mean}\left(\left\{R_k\left(\boldsymbol{x}_0^j, \boldsymbol{c}\right)\right\}_{j=1}^G\right)}{\sigma_{\max, k}}
\end{equation}

\subsubsection{GRPO Experiment Settings}
\label{Apdx:grpo-expr}
\begin{table}[h]
\centering
\caption{GRPO Experiment Settings}
\small
\begin{tabular}{llll}
\specialrule{1.5pt}{0pt}{0pt} 
\textbf{Parameter} & \textbf{Value} & \textbf{Parameter} & \textbf{Value} \\
\midrule
Group size & 4 & \# Sampling steps & 16 \\
Prompts per update & 64 & Timeshift & 12 \\
SDE steps range & [0, 6] & CFG & 4 \\
Online training & True & Learning rate & 1e-4 \\
Policy loss weight & 1 & LoRA dim & 128 \\
KL loss weight & 3e-4 & LoRA alpha & 64 \\
HPSv3-general reward weight & 1 & LoRA layers & Linear layers in all Self-Attention, \\
HPSv3-percentile reward weight & 1 & & Cross-Attention, FFN layers \\
MQ reward weight & 1 & & \\
TA reward weight & 1 & & \\
\specialrule{1.5pt}{0pt}{0pt} 
\end{tabular}
\end{table}

\subsection{Appendix-B}
\label{Apdx:bsa}
\subsubsection{Modeling of Block Sparse Attention}

\paragraph{3D Block Rearrangement}

We consider a video sequence with shape $T \times H \times W$, stored in memory in the order $T, H, W$. This sequence is divided into $N_T \times N_H \times N_W$ 3D blocks, where $N_T = \lceil T/t \rceil$, $N_H = \lceil H/h \rceil$, and $N_W = \lceil W/w \rceil$, and each block has shape $t \times h \times w$. The blocks are arranged in memory in the order $[N_T, N_H, N_W]$ (block-wise order), and within each block, the elements are stored in the order $[t, h, w]$ (intra-block order). After this rearrangement, we obtain a reshaped sequence.

\paragraph{Block Selection Mask Construction}

Let $X$ be the input tensor after rearrangement. We compute the query $Q$ and key $K$ matrices using learnable weights $W_q$ and $W_k$:
\[
Q = X W_q \in \mathbb{R}^{b \times n_h \times s_q \times d}, \quad K = X W_k \in \mathbb{R}^{b \times n_h \times s_k \times d},
\]
where $b$ is the batch size, $n_h$ is the number of attention heads, $s_q$ and $s_k$ are the sequence lengths for queries and keys respectively (with  $s_q = s_k = T \times H \times W$  in this case), and $d$ is the feature dimension.

To reduce computational cost, we perform average pooling over each block. Let  $n = t \times h \times w$  be the number of elements per block. The pooled query $Q_{\text{pool}}$ and key $K_{\text{pool}}$ are computed by averaging over the elements within each block:
\[
Q_{\text{pool}}[:, :, b_q, :] = \frac{1}{n} \sum_{j=0}^{n-1} Q[:, :, (b_q-1)n + j, :] \quad \text{for} \quad b_q = 1, \ldots, N_q,
\]
\[
K_{\text{pool}}[:, :, b_k, :] = \frac{1}{n} \sum_{j=0}^{n-1} K[:, :, (b_k-1)n + j, :] \quad \text{for} \quad b_k = 1, \ldots, N_k,
\]
where $N_q = s_q / n$ and $N_k = s_k / n$ are the number of query and key blocks respectively.

The pooled score matrix $S_{\text{pool}}$ is then calculated as:
\[
S_{\text{pool}} = \frac{Q_{\text{pool}} K_{\text{pool}}^\top}{\sqrt{d}} \in \mathbb{R}^{b \times n_h \times N_q \times N_k},
\]
where $K_{\text{pool}}^\top$ denotes the transpose of the last two dimensions of $K_{\text{pool}}$.

For each query block $i \in [0, N_q-1]$, we select the top $r$ key blocks based on the highest scores in $S_{\text{pool}}[:, :, i, :]$. This allows us to construct a binary mask matrix $M \in \mathbb{R}^{b \times n_h \times s_q \times s_k}$ as follows:
\[
M[:, :, in : (i+1)n, jn : (j+1)n] =
\begin{cases} 
1 & \text{if key block } j \text{ is in the top-} r \text{ neighbors of query block } i \\
0 & \text{otherwise}
\end{cases},
\]

\paragraph{Attention with Block Selection Mask}

Finally, we compute the masked attention. The attention score matrix $S$ is:
\[
S = \frac{Q K^\top}{\sqrt{d}} \in \mathbb{R}^{b \times n_h \times s_q \times s_k},
\]
where $K^\top$ is the transpose of the last two dimensions of $K$. We then apply the mask:
\[
S_{\text{masked}} = 
\begin{cases} 
S & \text{where } M = 1 \\
-\infty & \text{where } M = 0 
\end{cases},
\]
and the attention weights are obtained by applying softmax along the last dimension:
\[
O = \text{softmax}(S_{\text{masked}}).
\]

\subsubsection{Modeling of Ring Block Sparse Attention for Context Parallelism}
We extend the sparse attention computation with context parallelism. Given a tensor parallelism size of $N_{cp}$, each parallel rank maintains a local segment of $\frac{T \times H \times W}{N_{cp}}$ latents. Let $Q_i, K_i, V_i \in \mathbb{R}^{b \times n_h \times \frac{T \times H \times W}{N_{cp}} \times d}$ denote the query, key, and value tensors respectively for the $i$-th rank.

\paragraph{Local Block Selection Mask Construction}
To compute the block-sparse attention mask $M_i \in \mathbb{R}^{b \times n_h \times \frac{N_q}{N_{cp}} \times N_k}$ for rank $i$, each rank first computes its own local pooled keys:
\[
K_{\text{pool}_j}[:, :, b_j, :] = \frac{1}{n} \sum_{m=0}^{n-1} K_j[:, :, (b_j-1)n + m, :] \quad \text{for} \quad b_j = 1, \ldots, \frac{s_k}{N_{cp}},
\]

where $K_j = K[:,:,(j-1) \frac{s_k}{N_{cp}}:j\frac{s_k}{N_{cp}},:]$, $j\in [1, N_{cp}]$. Then we gather the pooled key representations and compute the pooled score matrix for rank $i$:
\[
S_{\text{pool}_i} = \frac{Q_{\text{pool}_i} \left( \bigoplus_{j=1}^{N_{cp}} K_{\text{pool}_j} \right)^\top}{\sqrt{d}}
\]
where $\bigoplus$ denotes concatenation along the sequence dimension and $Q_i = Q[:,:,(i-1) \frac{s_q}{N_{cp}}:i\frac{s_q}{N_{cp}},:]$, $i\in [1, N_{cp}]$. Based on $S_{\text{pool}_i}$, the mask $M_i$ is constructed by selecting the top-$r$ key blocks for each query block across all batches and heads.

To optimize efficiency, we employ a ring-attention communication pattern where the computation of local pooled scores overlaps with the communication of $K_{\text{pool}_i}$ tensors between adjacent ranks.

\paragraph{Ring Attention with Local Block Selection Mask}
\label{Apdx:bsa-cp}
Once $M_i$ is obtained, each rank computes its attention output $O_i$ by the online softmax algorithm with $M_{ij} \in \mathbb{R}^{b \times n_h \times \frac{N_q}{N_{cp}} \times \frac{N_k}{N_{cp}}}$, which is the block of mask $M_i$ corresponding to rank $j$. Ring-attention \citep{liu2023ring} is adopted to overlap the attention computation and the communication of $K_j, V_j$.

\subsubsection{Implementation Details}

Our hardware-aligned 3D Block Sparse Attention operator is implemented using Triton\citep{tillet2019triton}, building upon the implementation of Flash Attention\citep{dao2023flashattention}. We implemented both forward and backward passes for both single-GPU and context-parallel configurations.  

\paragraph{3D block size}

The 3D block size is set to $t = h = w = 4$. This configuration represents a trade-off between speed and flexibility. In our implementation, the fastest performance is achieved when $t_q \cdot h_q \cdot w_q = 128$ and $t_k \cdot h_k \cdot w_k = 1024$ (i.e., the default configuration of $t \cdot h \cdot w = 64$ is not the fastest due to the hardware alignment), but this comes at the cost of reduced flexibility in handling varying resolutions, especially $N_{cp}$ is large. In our experiments, we observed no significant differences in post-training results across various tested configurations of 3D block sizes, with $t_q \cdot h_q \cdot w_q$ values in [64, 128] and $t_k \cdot h_k \cdot w_k$ values in [64, 128, 256, 512, 1024].

\paragraph{Sparsity}


The hyperparameter $r$ controls the number of key blocks selected per query block. The computational complexity scales linearly with $r$. We set $r$ to $\frac{1}{8} N_k$ during the distillation training phase and to $\frac{1}{16} N_k$ during the refinement-expert training phase.


\paragraph{Construction of the Block Selection Mask}
\label{Apdx:block-selection-mask}
Regarding the construction of the block selection mask, two primary strategies are explored: \\
1) Top-$r$ mode: As described earlier, this approach selects the top $r$ key blocks based on their pooled attention scores. \\ 
2) CDF-$p$ mode: This method selects key blocks in descending order of their pooled scores until the cumulative softmax of the scores reaches a threshold $p$.  

In our experiments, the CDF-$p$ mode yields better generation quality under high speedup ratios in a training-free setting. However, in trainable scenarios, it suffers from the time cost caused by different number of key blocks selected by the query blocks. Therefore, we adopted the top-$r$ approach for our trainable implementation.

\subsection{Appendix-C}

\begin{figure}[htbp]
  \centering
  \includegraphics[width=0.4\textwidth]{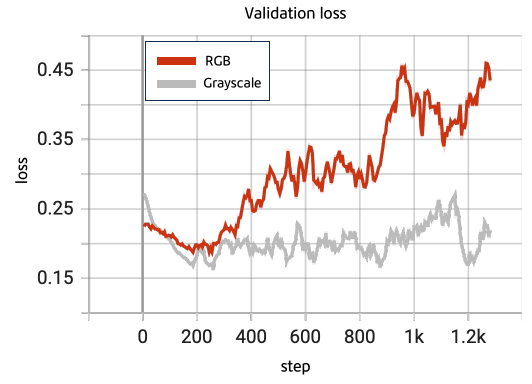}
  \caption{MQ Reward model validation loss curve}
  \label{fig:mq_rm_validation_loss}
\end{figure}